%% file: main.tex
\pdfoutput=1

\documentclass[11pt]{article}

\usepackage[final]{acl}

\usepackage{times}
\usepackage{latexsym}

\usepackage[T1]{fontenc}

\usepackage[utf8]{inputenc}

\usepackage{microtype}

\usepackage{inconsolata}

\usepackage{graphicx}
\usepackage{subcaption}
\usepackage{booktabs,siunitx}
\usepackage{color,soul}
\usepackage{amsmath}
\usepackage[bb=dsserif]{mathalpha}
\usepackage{bm}
\usepackage{multirow}
\usepackage{enumitem}
\setlist{nosep}

\usepackage{pifont}
\newcommand{\cmark}{\textcolor{teal}{\ding{51}}}%
\newcommand{\xmark}{\textcolor{red}{\ding{55}}}%
\newcommand{\lfrqa}{\textsc{Lfrqa}}
\newcommand{\rqa}{\textsc{Robustqa}}
\newcommand{\arena}{\textit{\textsc{Rag-qa Arena}}}

\newcommand{\mypar}[1]{\noindent{\textbf{#1\ }}}

\definecolor{aureolin}{rgb}{0.99, 0.99, 0.3}
\definecolor{pinegreen}{RGB}{116, 141, 95}

\title{RAG-QA Arena: Evaluating Domain Robustness for Long-Form Retrieval-Augmented Question Answering}

\author{
 \textbf{Rujun Han\textsuperscript{2\textdagger}} ~ 
 \textbf{Yuhao Zhang\textsuperscript{3\textdagger}} ~
 \textbf{Peng Qi\textsuperscript{4\textdagger}} ~
 \textbf{Yumo Xu\textsuperscript{1}} ~
 \textbf{Jenyuan Wang\textsuperscript{1}} ~ \\
 \textbf{Lan Liu\textsuperscript{1}} ~
 \textbf{William Yang Wang \textsuperscript{5\textdagger}}~
 \textbf{Bonan Min\textsuperscript{1}}~
 \textbf{Vittorio Castelli\textsuperscript{1}}~
\\
\\
 \textsuperscript{1}AWS AI Labs~
 \textsuperscript{2}Google~
 \textsuperscript{3}Samaya.ai~
 \textsuperscript{4}Orby.ai~
 \textsuperscript{5}University of California, Santa Barbara
\\
{\tt rujunh@google.com;  yuhao@samaya.ai; peng@orby.ai; william@cs.ucsb.edu} \\
{\tt \{liuall,yumomxu,jenwan,bonanmin,vittorca\}@amazon.com} \\
}

\begin{document}
\maketitle

\renewcommand*{\thefootnote}{\fnsymbol{footnote}}
\setcounter{footnote}{2}
\footnotetext{Work done at AWS AI Labs.}
\renewcommand*{\thefootnote}{\arabic{footnote}}
\setcounter{footnote}{0}

\input{00-abstract}
\input{01-intro}
\input{02-background}
\input{03-data}
\input{04-framework}
\input{05-experiments}
\input{05a-main-result-table}

\input{06-results}
\input{07-discuss}
\input{08-related}
\input{09-conclude}

\section*{Limitations}
We discuss some limitations of this work for future research efforts. \arena~can potentially cover more models. We didn't include some leading LLMs, such as Claude \cite{Claude3} and Gemini \cite{gemini} models, due to legal and resource constraints, but we plan to add them to the leaderboard in the future. Evaluation using \textsc{GPT-4-0125-preview} is not cheap. It costs on average 300 U.S. dollars per model on the full \lfrqa's test set. We plan to subsample 10-20\% of the queries for the final public leaderboard, which will be more cost-friendly for future users. Future research can also study training smaller but equally accurate models as evaluators. Finally, we mainly focus on different LLMs for RAG-QA in this work, but future research can study the impact of different retrievers or joint retrievers and LLM training using \arena.

\section*{Ethics Statement}
The authors of this paper are committed to conducting research ethically. We are leveraging existing LLMs to generate answers for \lfrqa, which include many open-ended questions. LLM-generated answers could be incorrect or unfaithful, as retrievers could find irrelevant passages and LLM can hallucinate \cite{Huang2023ASO}. These are known issues in the AI research community, and that is the reason we created \lfrqa~to better evaluate RAG-QA systems. The additional risks and potential harms are discussed in numerous previous works \cite{bender2021dangers, weidinger2021ethical}. The authors strive to ensure that the research and its results do not cause harm.

Data used in this work have been collected from public sources and used in accordance with all applicable laws and regulations. We use contracted data professionals for \lfrqa~annotations, and Appen platform\footnote{https://www.appen.com/} for human pairwise preference annotations. In both cases, we ensure our hourly rate is higher than 15 U.S. dollars per local minimum wage standard. The intended usage of \lfrqa~is compatible with the underlying data's access conditions (Appendix~\ref{sec:license})

\bibliography{custom}

\appendix
\input{99-appendix}

\end{document}

%% file: 00-abstract.tex
\begin{abstract}

Question answering based on retrieval-augmented generation (RAG-QA) is an important research topic in NLP and has a wide range of real-world applications. However, most existing datasets for this task are either constructed using a single source corpus or consist of short extractive answers, which fall short of evaluating large language model (LLM) based RAG-QA systems on cross-domain generalization. To address these limitations, we create Long-form RobustQA (\lfrqa), a new dataset comprising human-written long-form answers that integrate short extractive answers from multiple documents into a single, coherent narrative, covering 26K queries and large corpora across seven different domains. We further propose \arena~by directly comparing model-generated answers against \lfrqa's answers using LLMs as evaluators. We show via extensive experiments that \arena\ evaluation and human judgments on answer quality are highly correlated. Moreover, only 41.3\% of the most competitive LLM's answers are preferred to \lfrqa's answers, demonstrating \arena~as a challenging evaluation platform for future research.\footnote{Code: \url{https://github.com/awslabs/rag-qa-arena}}

\end{abstract}

%% file: 01-intro.tex
\section{Introduction}
\label{sec:intro}

\begin{figure}[t]
    \centering
\includegraphics[trim={0cm 0.5cm 0cm 0.5cm}, angle=-90, clip, width=0.99\columnwidth]{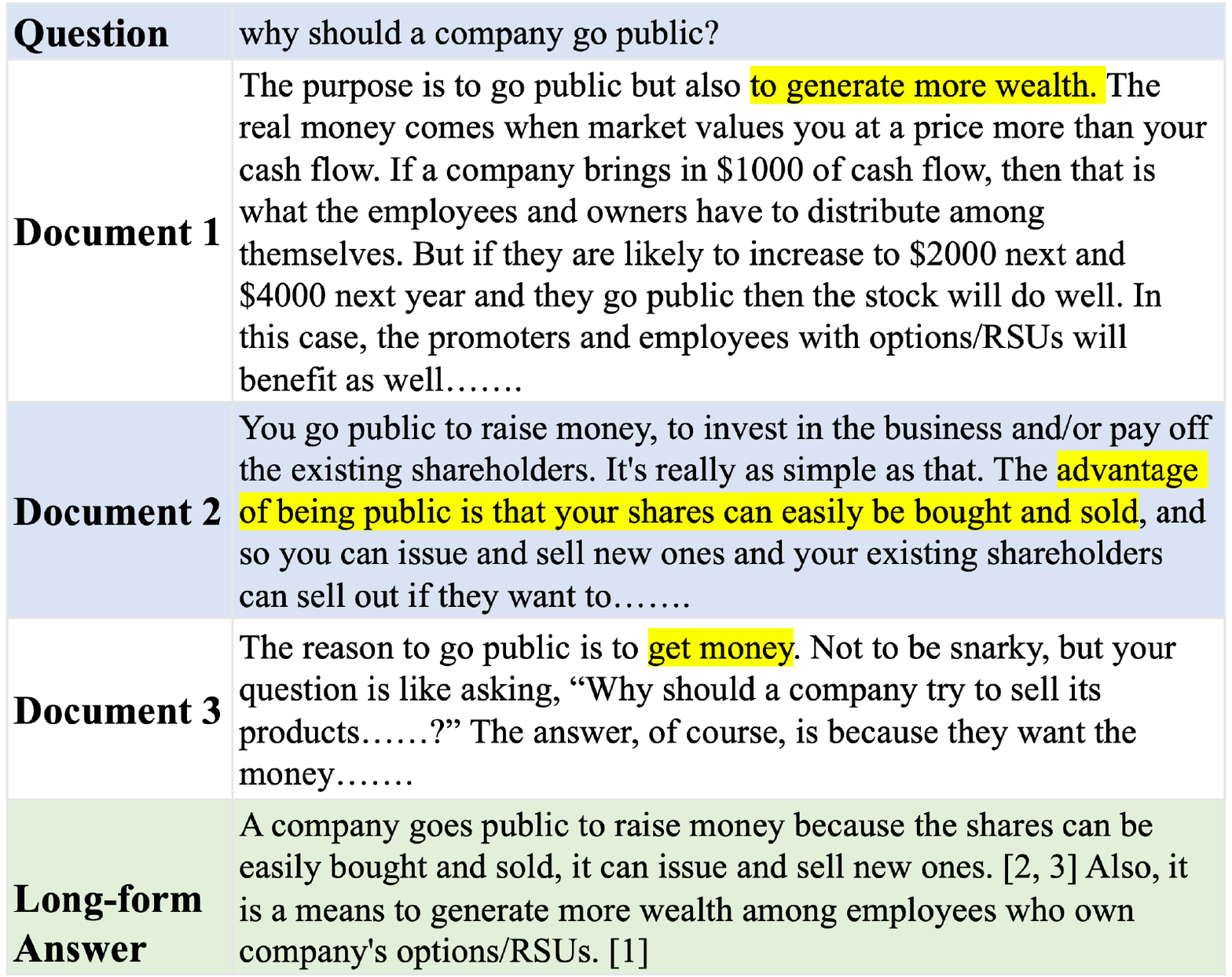}
\vspace{-0.2cm}
\caption{\lfrqa~annotation example. There are three documents (some text removed for brevity) relevant to the query. We instruct annotators to combine \rqa's \colorbox{aureolin}{answers} into a coherent long-form answer with added text if necessary. Citations [1], [2] and [3] indicate the supporting documents of each sentence.}
\label{fig:intro}
\vspace{-0.5cm}
\end{figure}

\begin{table*}[h]
\centering
\small
\scalebox{0.9}{
\setlength{\tabcolsep}{5pt}
\begin{tabular}{l|c|c|c|c|c|c|c}
\toprule
\textbf{Dataset} & Answers grounded & Long-form & Multiple & Coherent & Multiple & Human & \# Test\\
\textbf{Name} & in corpus & answers & documents & answers & domains & annotated & queries\\
\midrule
\lfrqa & \cmark & \cmark & \cmark & \cmark & \cmark & \cmark & 16.1K \\
\midrule
\rqa \cite{han-etal-2023-robustqa} &\cmark & \xmark & \cmark & \xmark & \cmark & \cmark & 16.1K \\
\textsc{NQ} \cite{kwiatkowski-etal-2019-natural} &\cmark & \xmark & \xmark & \xmark & \xmark & \cmark & \phantom{0}3.6K \\
\textsc{MultiHop-RAG} \cite{MultiHop-RAG} & \cmark & \xmark & \cmark & \cmark & \cmark & \cmark & \phantom{0}2.5K\\
\textsc{ASQA} \cite{stelmakh-etal-2022-asqa} & \cmark & \cmark & \cmark & \cmark & \xmark & \cmark & \phantom{0}1.0K \\
\textsc{LongFact} \cite{LongFact} & \xmark & \cmark & \xmark & \cmark & \cmark & \xmark & \phantom{0}2.3K \\
\textsc{ELI5} \cite{fan-etal-2019-eli5} & \xmark & \cmark & \cmark & \cmark & \xmark & \cmark & 25.0K \\
\bottomrule
\end{tabular}}
\vspace{-0.3cm}
\caption{Comparison of datasets. \lfrqa~distinguishes from previous work by uniquely encompassing seven features: 1) RAG-QA dataset with answers annotated based on underlying corpus; 2) Long-form answers of paragraph length; 3) Multiple documents that provide different facts/views; 4) Coherent answers that handle conflicting information; 5) Multiple-domain corpus to benchmark domain robustness; 6) Human annotated high-quality answers; 7) Large-scale evaluation set.}
\label{tab:data-compare}
\vspace{-0.6cm}
\end{table*}

Traditional reading comprehension task is constrained to fixed contexts \cite{rajpurkar-etal-2016-squad, kocisky-etal-2018-narrativeqa, huang-etal-2019-cosmos}. It is inadequate at addressing real-world questions, where no context is readily provided for a system to find answers. Such open-ended questions require a system to identify answers in an enormous knowledge base (e.g., Wikipedia) that is computationally prohibitive to feed into question answering systems such as large language models (LLMs). Retrieval-augmented generative question answering (RAG-QA) becomes an effective tool to filter out massive amounts of noise and select only a few highly relevant passages for LLM-based QA models.

The wide applications of RAG-QA \cite{Gao2023RetrievalAugmentedGF} necessitate the evaluation of systems' out-of-domain (OOD) performances, because a real-world system often confronts new data unseen during training. Existing popular benchmark datasets such as Natural Questions (NQ) \cite{kwiatkowski-etal-2019-natural} and TriviaQA \cite{joshi-etal-2017-triviaqa} consist solely of Wikipedia or Web documents, which fall short at measuring OOD performances.

\rqa~\cite{han-etal-2023-robustqa} was the first dataset created to benchmark cross-domain robustness for RAG-QA. However, as illustrated by the yellow highlights in Figure~\ref{fig:intro}, \rqa~follows NQ's annotation format with short answer spans extracted from the documents. Such data format is not the most suitable reference answer to evaluate the current leading LLMs that typically generate long-form responses with multiple pieces of information combined in one coherent narrative \cite{GPT3, InstructGPT, GPT4}. Consequently, token overlap metrics used in the extractive QA era \cite{karpukhin-etal-2020-dense, atlas} will penalize unfairly on the additional supporting tokens generated by LLMs, resulting in extremely low overlap scores. As an example, Fig.~\ref{fig:intro}'s extractive \colorbox{aureolin}{answers} have poor Exact-Match or $F_1$ scores with the final long-form answer. To create a long reference answer, one could simply concatenate these short answers, but the synthesized answers are either incoherent or ill-formatted, as illustrated by examples in Sec.~\ref{sec:annotation}. 

To address these drawbacks, we propose long-form RobustQA (\lfrqa) that integrates multiple short extractive answers into a coherent long-form answer. Figure~\ref{fig:intro} shows an annotation where three extractive answers are combined by annotators to create a comprehensive answer. Table~\ref{tab:data-compare} summarizes seven features in \lfrqa~ that make it uniquely beneficial for RAG-QA evaluations. \textsc{ASQA} \cite{stelmakh-etal-2022-asqa} and \textsc{ELI5} \cite{fan-etal-2019-eli5} are the most similar datasets to \lfrqa. However, they are either not directly annotated against the underlying corpus (thus, not RAG-QA), or rely on single-domain data, which is insufficient to benchmark systems' cross-domain performances.

With \lfrqa~annotations, we propose \arena~that leverages model-based evaluators to directly compare LLMs' answers with \lfrqa~without the necessity to examine long and potentially noisy retrieved passages. By demonstrating the high correlation with human judges following the same instruction and rubrics, we show that \arena~is an efficient and accurate framework to benchmark the RAG-QA system's cross-domain performances. In this work, we primarily focus on the LLMs used for the QA component, but \arena~can be easily extended to study retrieval's impact on answer generation quality.

We summarize our contributions: 1) We present \lfrqa, the first high-quality and large-scale multi-domain human annotations with coherent long-form answers for RAG-QA. 2) We propose an efficient model-based evaluation framework, \arena~that enables users to directly compare LLMs' answers with ground-truth answers in \lfrqa. 3) We build a dashboard incorporating a wide range of leading LLMs and conduct in-depth analysis to show that \lfrqa's answers are preferred significantly more to the best LLMs with long context. Therefore, we believe \arena~will serve as a challenging and robust evaluation benchmark for future RAG-QA research.

\begin{figure*}[t]
    \centering
\includegraphics[trim={6.8cm 0cm 5.5cm 0cm}, angle=-90, clip, width=0.99\textwidth]{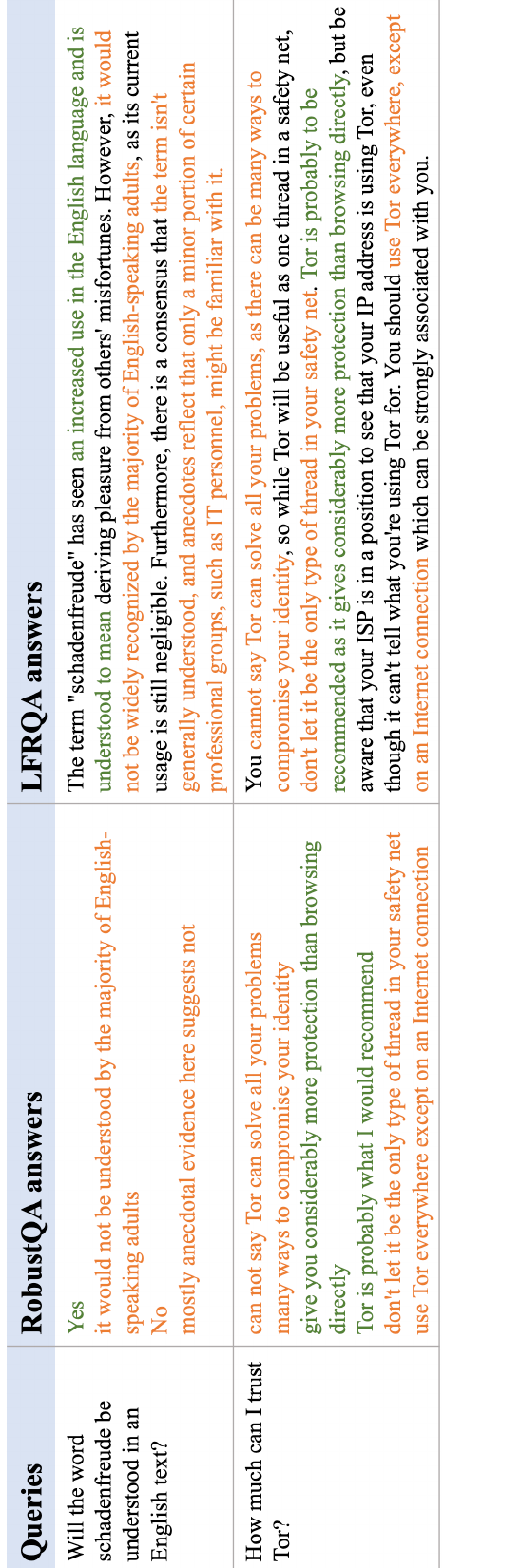}
\vspace{-1cm}
\caption{\lfrqa~v.s. \rqa. Citations are removed in \lfrqa's answers, and a few answer spans are removed for clarity. \textcolor{pinegreen}{Green} and \textcolor{orange}{orange} texts represent positive and negative opinions, respectively.}
\label{fig:examples}
\vspace{-0.6cm}
\end{figure*}

%% file: 02-background.tex
\section{RAG-QA Task Formulation}
\label{sec:background}

We briefly introduce the RAG-QA task in this section. Passage retrieval is the first step of a RAG-QA pipeline. Following the passage retrieval set-up in DPR \cite{karpukhin-etal-2020-dense} and \rqa ~\cite{han-etal-2023-robustqa}, we denote a collection of documents as $\mathcal{D}$. We split each document $d^i \in \mathcal{D}$ with a fixed length $N$ tokens and obtain a collection of $M$ ($\ge |\mathcal{D}|$) passages denoted as $\mathcal{C} = \{p_1, p_2, ... p_m, ... p_M\}$, where $p_m$ is a passage. Given a question $q$, the passage retrieval task is to select $K$ most relevant passages for $q$ with a retriever $\mathcal{R}$ from $\mathcal{C}$. Formally, $\mathcal{R}(q, \mathcal{C}) \rightarrow \mathcal{C}_q$. 

Upon receiving top $K$ passages or $\mathcal{C}_q$, a QA model reads them as context to generate an answer for the query. Unlike the extractive QA setting in \rqa~and NQ, we adopt \textit{generative QA} as it is most compatible with the generative nature of the leading LLMs with the flexibility to produce free-form answers. The answer generation task can be modeled as $\Sigma_1^T \mathcal{P}_q(w_t | w_{0:t-1}; \mathcal{C}_q)$, where $\mathcal{P}$ is an LLM. We focus on the variations of $\mathcal{P}$ and fix $\mathcal{R}$ in this work.

In real-world applications, we deploy RAG-QA systems into various domains such as healthcare, finance, and technology whose corpus and query types may not be well covered in a trained retriever and LLM. \citet{lewis-etal-2021-question} and \citet{han-etal-2023-robustqa} show significant performance gaps between in-domain and out-of-domain data for RAG-QA systems, further verifying domain adaptation problems. Therefore, it is crucial to gauge the domain robustness of RAG-QA based on LLMs, and \lfrqa~helps achieve this evaluation goal.  

%% file: 03-data.tex
\section{Data Creation}
\label{sec:data}

\begin{table*}
\centering
\small
\begin{tabular}{l|l|l|rrr|rr|rr}
\toprule
 & & & \multicolumn{3}{c|}{\textbf{Test Set}} & \multicolumn{2}{c}{\rqa} & \multicolumn{2}{|c}{\lfrqa} \\
\midrule
 \textbf{Domain} & \textbf{Source} & \textbf{Label} & \multicolumn{1}{c}{\textbf{$|Q|$}}  & \multicolumn{1}{c}{\textbf{$|D|$}} & \multicolumn{1}{c}{\textbf{$|P|$}}  & \multicolumn{1}{|c}{$A/Q$} & \multicolumn{1}{c}{\textbf{$W/A$}} & \multicolumn{1}{|c}{\textbf{$A/Q$}} & \multicolumn{1}{c}{\textbf{$W/A$}}  \\
\midrule
Biomedical & BioASQ & [BI] & 1,956 & 15,559,026 & 37,406,880 & 2.6 & 2.4 & 1.0 & 30.0 \\
Finance & FiQA  & [FI] & 3,612 & 57,638 & 105,777 & 3.0 & 9.4 & 1.0 & 69.1 \\
Lifestyle & LoTTE & [LI] & 2,208 & 119,461 & 241,780 & 5.7 & 8.7 & 1.0 & 99.5 \\
Recreation & LoTTE & [RE] & 2,094 & 166,975 & 315,203 & 3.2 & 7.2 & 1.0 & 60.3 \\
Technology & LoTTE & [TE] & 2,111 & 638,509 & 1,252,402 & 6.0 & 8.7 & 1.0 & 99.7 \\
Science & LoTTE & [SC] & 1,423 & 1,694,164 & 3,063,916 & 5.3 & 7.8 & 1.0 & 92.0 \\
Writing & LoTTE & [WR] & 2,695 & 199,994 & 347,322 & 6.2 & 6.6 & 1.0 & 88.0 \\
\bottomrule
\end{tabular}
\vspace{-0.2cm}
\caption{Data (test set) summary: \lfrqa~v.s. \rqa. $|Q|$, $|D|$, $|P|$, $A/Q$, and $W/A$ represent numbers of questions, documents, passages, answers per question, and words per answer, respectively. Each passage consist of 100 words at most. \lfrqa~ has only one answer per query as we integrate multiple answers from \rqa, which results in more words in (long-form) answers. Dev set statistics can be found in Appendix Table~\ref{tab:data-stats-dev}.}
\label{tab:data-stats}
\vspace{-0.3cm}
\end{table*}

\lfrqa~consists of two types of QA samples: 1) new annotations in Finance ([FI]), Lifestyle ([LI]), Recreation ([RE]), Technology ([TE]), Science ([SI]), and Writing ([WR]) domains; 2) adapted long-form BioASQ ([BI]). We describe the details of both QA samples in the following sections.

\subsection{Annotated Data}
\label{sec:annotation}

Following \rqa~\cite{han-etal-2023-robustqa}, \lfrqa's new annotations are also based on the LoTTE and FiQA queries and corpus. \textsc{LoTTE} was proposed in the ColBERTv2 paper \cite{santhanam-etal-2022-colbertv2} and consists of information retrieval (IR) datasets across five domains: lifestyle, recreation, technology, writing, and science, each can have relevant answers coming from either web search or on-line forum. \textsc{FiQA} \cite{fiqa} proposes a task, ``Opinion-based QA over financial data'' that answers finance-related questions from financial corpora such as microblogs, reports, and news. It is important to note that both FiQA and LoTTE are IR datasets with answers as long documents, which may include a large amount of irrelevant information to the query. 

As IR datasets, both FiQA and LoTTE could only provide relevant documents to users, as there are no precise answer annotations. \rqa~addresses this short-coming by extracting short answer spans from the long documents in the similar format of NQ, which serves as a high-quality benchmark for extractive RAG-QA. Figure~\ref{fig:intro} shows an example where the yellow highlights in Documents 1-3 are the extracted answers to the question. 

\mypar{Limitations of extractive RAG-QA.} In the era of LLMs, models' responses to user queries are often long and comprehensive \cite{GPT4, Claude3, llama-3, mixtral}, which the short, extractive reference answers in \rqa~are no longer the most compatible format to evaluate against. First, in \rqa, annotators are limited to only taking 3 answer spans per relevant document, each with no more than 16 words. This process could result in a loss of useful information to help answer the query. Second, to reconcile multiple extractive references for model evaluation, prior work in extractive RAG-QA \cite{karpukhin-etal-2020-dense, han-etal-2023-robustqa, atlas} adopt the maximum of token overlaps between a model prediction and a list of references to compute EM or $F_1$ metrics, which penalizes unfairly the long-form responses from modern LLMs. Finally, if we naively concatenate or list all short answer spans as shown by examples in Figure~\ref{fig:examples}, the combined answers are often too ill-formatted or incoherent as ground-truth answers.

\lfrqa~addresses all of these drawbacks by instructing the annotators to integrate all short answers in \rqa~into a coherent long-form answer. Below, we show a summary of our annotation instruction and quality control mechanism.

\mypar{Annotation instruction.} As Fig.~\ref{fig:intro} shows, a query, all relevant documents, and original short answers (highlighted in the documents) are presented to annotators on a single annotation page. Annotators need to combine all highlighted answers into a single complete and coherent answer. All highlighted answers MUST be included; otherwise, the annotation is considered as a failure. Annotators are encouraged to include more information in the documents if it helps to answer the queries. To ensure annotators faithfully use the document information, we request annotators to provide citations after each answer sentence. For example, the first sentence in Fig.~\ref{fig:intro} is composed using information from Documents 2 and 3. Annotators should add "[2, 3]" after that sentence. We use these citations primarily for data quality control and remove them during the answer evaluation. The actual annotation UI can be found in Appendix Fig.~\ref{fig:anno-ui}.

\mypar{Quality control.} The data annotations are performed by contracted data professionals. We also have a dedicated team of data linguists to validate the annotation quality. Specifically, our data linguists randomly audit 10\% of each batch of the annotations, and if the valid answer ratio is $< 90\%$, we send the batch back to the annotators for re-work. The process iterates until the valid answer ratio exceeds $90\%$. Here is a list of failure cases:
\mypar{1. Incompleteness:} Answers do not include all highlighted answers, or there is clear relevant information in the documents, but not included in the answer.
\mypar{2. Redundancy:} Clear irrelevant information is included in the answer.
\mypar{3. Incoherence:} Answers are not coherent or not written in natural English.
\mypar{4. Citation Error:} Wrong/missing citations, which indicate annotators do not use correct information from the right documents.

\subsection{Adapted Data}
For the biomedical domain, \lfrqa~leverages the same set of test queries as in \rqa, but uses the complete rather than span answers in the BioASQ dataset. The original BioASQ annotations provide two types of answer formats: 1) exact answer, which is the short extractive answers used in \rqa; 2) ideal answer, which is a long-form abstractive answer to be consistent with other datasets in this work. We did not perform further annotations. We notice in Table~\ref{tab:data-stats} that BioASQ's answers are shorter compared with other datasets. This is due to its dominant amount of factoid queries, which do not require elaborated explanations as in other datasets with more open-ended reasoning questions \cite{han-etal-2023-robustqa}.

We drop SearchQA in \rqa~as this dataset only has short-form extractive answers, and its documents contain a significant amount of text omission (``...'') that prevents us from re-constructing long-form answers.

\subsection{Data Statistics and Analysis}
\label{sec:analysis}

Table~\ref{tab:data-stats} summarizes the statistics for the test set, which consists of 16K queries across 7 domains. We filter out queries with more than 80 ground-truth documents, resulting in 73 fewer queries compared with \rqa. Since \lfrqa~combines multiple short answers, the answer per query ratio (A/Q) is always 1, and the word per answer ratio (W/A) is substantially higher compared with \rqa. We also annotate a dev set with 10K queries for future model development purposes, and the statistics can be found in Appendix Table~\ref{tab:data-stats-dev}. We conduct further analysis below to demonstrate the unique contributions of \lfrqa~in Table~\ref{tab:data-compare}.

\begin{figure}[t]
\centering
    \begin{subfigure}[b]{0.45\columnwidth}
    \includegraphics[trim=1cm 0.5cm 1cm 0cm, width=0.95\columnwidth]{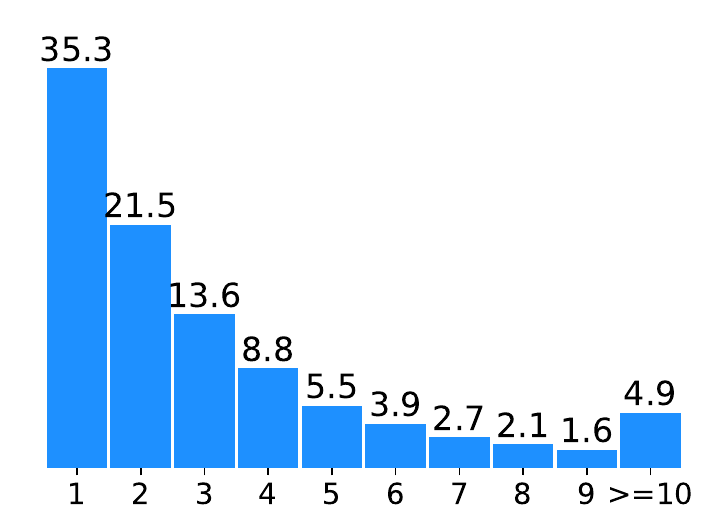}
    \caption{\# of documents used per answer.}
    \label{fig:doc-dist-1} 
    \end{subfigure}
    ~
    \begin{subfigure}[b]{0.45\columnwidth}
    \centering
    \includegraphics[trim=1cm 0.5cm 1cm 0cm, width=0.95\columnwidth]{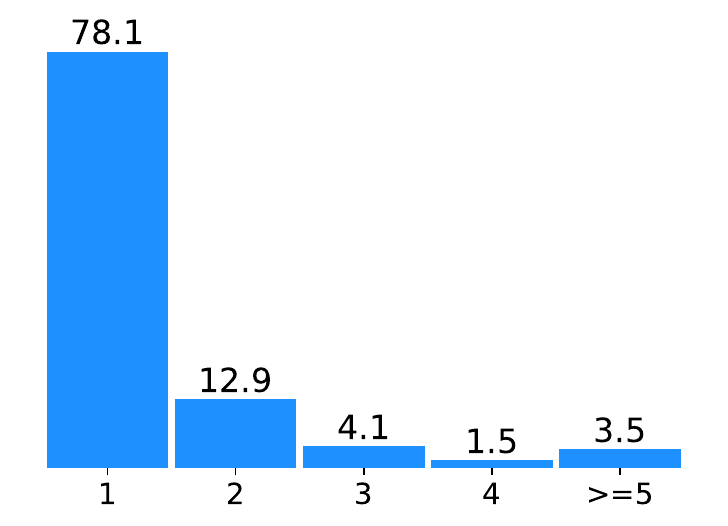}
    \caption{\# of documents used per answer sentence.}
    \label{fig:doc-dist-2} 
    \end{subfigure}
\vspace{-0.2cm}
\caption{Distribution of number (\#) of documents used in \lfrqa's answers. All numbers are \%.}
\vspace{-0.6cm}
\label{fig:doc-dist} 
\end{figure}

\mypar{Answers over multiple documents.} Figure~\ref{fig:doc-dist} illustrates the distribution of number of documents used by \lfrqa's answers. Specifically, Figure~\ref{fig:doc-dist-1} shows that around 65\% of the answers use $\ge$ 2 documents' information. 4.9\% of the answers consist of information from 10 or more documents (maximum = 80). In Figure~\ref{fig:doc-dist-2}, we divide long-form answers into sentences and show the distribution of the number of documents used per answer sentence. Nearly 22\% of the answer sentences combine information from multiple documents. Both show that \lfrqa's answers effectively combine information across multiple ground-truth documents. This makes \lfrqa~ challenging for RAG-QA, as it requires identification and aggregation of information across sources.

\mypar{Coherent answers.} \lfrqa's answers further organize facts and views across multiple documents in a coherent paragraph. Answers with multiple views are common in the original \rqa's answers. Conducting a string match of both ``yes'' and ``no'' as a leading word in an answer list, we found more than 200 examples with such conflicting information. This does not account for more subtle cases where answers semantically contradict each other. In Figure~\ref{fig:examples}, we show 2 examples with conflicting views. \rqa's annotations simply list them (separated by new-lines), whereas \lfrqa's answers organize them as coherent narratives, with conflicting information reconciled in helpful context. 

\mypar{Fluency.} \rqa's answers are extracted from documents and often cut off unnaturally to satisfy the limit of 16 words. In Figure~\ref{fig:examples}, \rqa's answers, such as ``many ways to compromise your identity'' and ``mostly anecdotal evidence here suggest no'' are incomprehensible without further context, whereas \lfrqa's answers are all well written in complete sentences. 

All of these features show that \lfrqa~provides both challenging and high-quality annotations for evaluating RAG-QA systems.

%% file: 04-framework.tex
\section{RAG-QA Arena}
\label{sec:eval}

\begin{figure}[t]
    \centering
\includegraphics[trim={8cm 0cm 8cm 0cm}, clip, angle=-90, width=0.95\columnwidth]{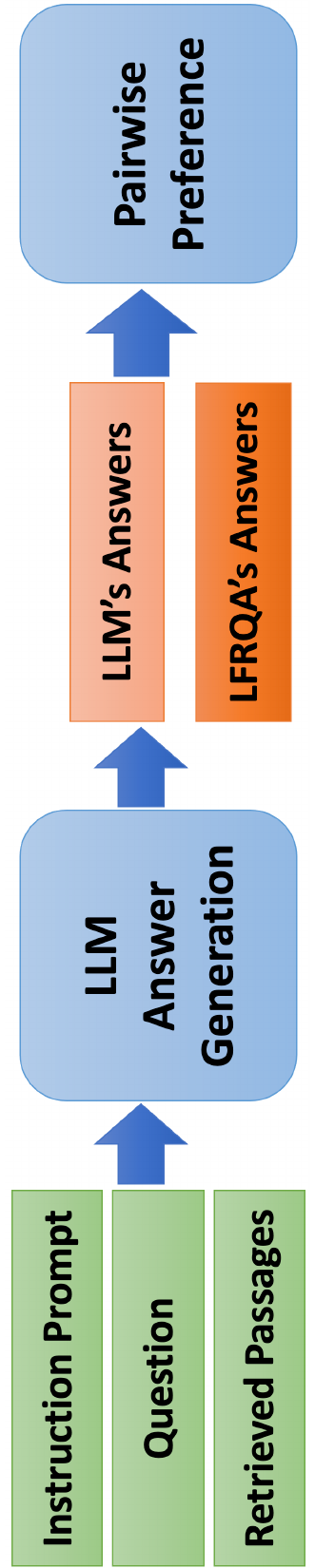}
\vspace{-0.2cm}
\caption{\arena~framework. Green blocks are LLM's inputs to generate answers. Orange blocks are LLM and LFRQA's answers presented to both human and LLM judges to determine pairwise preferences.}
\label{fig:eval-framework}
\vspace{-0.3cm}
\end{figure}

In this section, we propose our evaluation framework \arena. Inspired by the pairwise human preference evaluation framework such as Chatbot Arena \cite{Chiang2024ChatbotAA}, we calculate win-rate and win+tie rate against \lfrqa~as ground-truth as a metric to gauge systems' RAG-QA quality. Figure~\ref{fig:eval-framework} illustrates the evaluation framework.

The choice of \lfrqa~as the target to compare has been partly justified in Sec.~\ref{sec:data} for 1) Completeness: the annotation process encourages the inclusion of as much relevant information as possible. 2) Coherence: its answers are written more coherently and naturally than \textsc{Robust-QA} as references for LLM generations. Complete and coherent answers can be considered as a comprehensive summary of all relevant information in the entire corpus. This allows us to evaluate generated answers against \lfrqa~answers only, which is much more informative and concise than using retrieved passages, potentially with a large amount of noise.

We implement human and model-based evaluations with the same instructions and report their correlations. We will show results in Sec.~\ref{sec:lfrqa-rqa} that further justify using \lfrqa~as evaluation targets.

\subsection{Human Evaluation}
We present a query and a pair of answers (one from \lfrqa~and one from an LLM), to human annotators. We instruct them to rate their preferences based on three aspects. \mypar{1. Helpfulness:} information that is helpful/relevant to answer the query. \cite{llama-2, Bai2022TrainingAH}.
\mypar{2. Truthfulness:} information that is correct to answer the query. By our definition, truthful information should also be helpful information \cite{truthfulqa, truthfuleval}.
\mypar{3. Completeness:} include as much truthful and helpful information as possible. We further instruct annotators to use Truthfulness (being both truthful and helpful) as the primary criterion since it is stricter than Helpfulness. Helpfulness is used when a decision cannot be made by Truthfulness alone. More details including the definition of aspects, rating categories, and step-by-step guidelines can be found in Appendix~\ref{sec:metric-define} and Fig.~\ref{fig:eval-ui-a}-\ref{fig:eval-ui-b} (annotation interface).

\subsection{Model-Based Evaluation}

Since human evaluation is too costly, we adopt model-based evaluators for scalable evaluation of LLMs on the entire \lfrqa~test set.

As for the evaluation approach, we provide LLM-based evaluators with a query and a pair of answers (including one from \lfrqa). Similar to human evaluation, we prompt LLMs to rate their preferences based on the same three aspects above. We only modify the human instruction slightly to be compatible with LLM readable input text, but the majority of the prompt, especially the input data, and rubric, stay the same (Appendix Table~\ref{tab:pair-eval-instruction}-\ref{tab:pair-eval-examples}).

For both human and model-based evaluations, we allow ``tie" (no preference) as an option. For human evaluation, we take the majority votes from 3 annotators to mitigate biases. If there is no majority vote, we default the label to ``tie.''

%% file: 05-experiments.tex
\section{Experimental Setup}
In this section, we discuss our retriever, LLMs experimented for both answer generation and pairwise evaluations and their prompts in more detail.

\mypar{Retrieval setting.} We employ \textsc{ColBERTv2} \cite{santhanam-etal-2022-colbertv2} as our passage retriever, considering its superior performance on the underlying corpus for both \rqa~and \lfrqa~as shown in \citet{han-etal-2023-robustqa}. We follow the same retrieval setting and split passages into text chunks with 100 consecutive words. We use the top 5 retrieved passages for our main results in Table~\ref{tab:eval-mbe-all} and experiment with the top 10 passages for further analysis.

\mypar{Answer generation.} We consider LLMs ranked top 25\footnote{Ranking at the time of paper writing.} in the Chatbot Arena \cite{Chiang2024ChatbotAA} and their smaller version to show the impact of model sizes. Due to resource and legal constraints, for proprietary LLMs, we only use OpenAI models: a) \textsc{GPT-4-turbo} (2024-04-09), b) \textsc{GPT-4o} and c) \textsc{GPT-4-0125-preview}). For public models, we experiment with 1) \textsc{Mixtral-8x22B-Instruct} and  \textsc{Mixtral-8x7B-Instruct} \cite{mixtral}; 2) \textsc{Llama-3-70b-Instruct} and \textsc{Llama-3-8b-Instruct} \cite{llama-3}; 3) \textsc{Command R+} and \textsc{Command R} \cite{commandr}; 4) \textsc{Qwen1.5-110b-chat} and \textsc{Qwen1.5-32b-chat} \cite{qwen}. Answer generation prompt can be found in Appendix Table~\ref{tab:answer-gen}.

\mypar{Pairwise evaluation.} For LLM-based evaluators, we focus on a few larger models with strong context understanding capability, such as \textsc{GPT-4-turbo}, \textsc{GPT-4-0125-preview}, \textsc{GPT-4o}, \textsc{Mixtral-8x22B-Instruct} and \textsc{Llama-3-70b-Instruct}. Appendix Table~\ref{tab:pair-eval-instruction}-\ref{tab:pair-eval-examples} show the details of the pairwise evaluation prompts, including instruction, example prompt template, and in-context-learning examples. We shuffle the order of the answer pairs so that both human and model judges are not biased by the position of an answer. We select the LLM with the highest correlation with human judgments as the evaluator (Appendix Table~\ref{tab:eval-model-select}). We use OpenAI API to run GPT-4 models. We download public models from HuggingFace Hub \cite{Huggingface} and run them on up to 8 Nvidia A100 GPUs with PyTorch (1.13.0) and Transformers (4.41.0) whose \texttt{tokenizer.apply\_chat\_template()} function can help adapt the generic prompts to different LLMs' input formats.

We follow OpenAI's recommendation\footnote{https://platform.openai.com/docs/guides/prompt-engineering} to design prompts with chain-of-thoughts (CoT) \cite{CoT}, in-context learning \cite{icl}, and HTML tags as delimiters. We remove the thinking process in model outputs as final answers.

%% file: 05a-main-result-table.tex
\begin{table*}[t]
\centering
\small
\scalebox{0.82}{
\setlength{\tabcolsep}{4.5pt}
\begin{tabular}{l|cc|cc|cc|cc|cc|cc|cc|cc}
\toprule

 & \multicolumn{2}{c|}{\textbf{Overall}} & \multicolumn{2}{c|}{[BI]} & \multicolumn{2}{c|}{[FI]} & \multicolumn{2}{c|}{[LI]} & \multicolumn{2}{c|}{[RE]} & \multicolumn{2}{c|}{[TE]} & \multicolumn{2}{c|}{[SC]} & \multicolumn{2}{c}{[WR]} \\
\midrule
\textbf{Compared Models} & W & W+T & W & W+T & W & W+T & W & W+T & W & W+T & W & W+T & W & W+T & W & W+T \\
\midrule
\textsc{GPT-4o}* \textsuperscript{\#1} & \textbf{36.9} & \textbf{41.0} & \textbf{52.9} & \textbf{59.3} & 38.4 & 42.3 & \textbf{25.1} & \textbf{27.9} & \textbf{40.4} & \textbf{46.4} & 35.6 & 38.8 & \underline{42.8} & 47.6 & \textbf{28.4} & \textbf{31.1} \\
\textsc{GPT-4-turbo} \textsuperscript{\#2}  & 34.4 & \underline{39.1} & 36.0 & 43.9 & 40.6 & 45.1 & \underline{23.2} & \underline{26.1} & \underline{36.7} & \underline{44.1} & \textbf{36.6} & \textbf{39.6} & 42.6 & \underline{47.9} & \underline{26.2} & \underline{29.6} \\
\textsc{GPT-4-0125-preview} \textsuperscript{\#6}  & 28.9 & 33.7 & 31.4 & 40.1 & 36.8 & 40.8 & 18.1 & 21.3 & 31.5 & 38.6 & 30.4 & 34.0 & 34.7 & 40.5 & 19.2 & 22.3 \\

\midrule
\textsc{Mixtral-8x22b} \textsuperscript{\#3} & \underline{34.5} & 38.8 & 37.0 & 46.0 & \textbf{44.1} & \textbf{47.6} & 21.3 & 24.4 & 34.4 & 41.0 & 33.9 & 36.8 & \textbf{45.0} & \textbf{49.5} & 25.9 & 28.1 \\
\textsc{Mixtral-8x7b} \textsuperscript{\#7} & 27.5 & 31.0 & 31.9 & 39.1 & 35.3 & 38.5 & 15.9 & 18.4 & 24.8 & 29.5 & 30.3 & 32.1 & 33.9 & 38.0 & 20.0 & 21.9 \\
\midrule
\textsc{Llama-3-70b} \textsuperscript{\#8} & 21.7 & 25.2 & 30.3 & 37.2 & 24.6 & 27.7 & 12.9 & 15.1 & 22.3 & 27.3 & 22.4 & 24.4 & 25.6 & 30.0 & 15.5 & 18.2 \\
\textsc{Llama-3-8b} \textsuperscript{\#10} & 20.4 & 23.5 & 34.7 & 39.6 & 24.0 & 27.0 & 11.2 & 13.2 & 19.4 & 24.7 & 20.5 & 22.5 & 22.3 & 26.1 & 12.5 & 14.4 \\
\midrule
\textsc{Command R+} \textsuperscript{\#9} & 21.1 & 25.8 & 26.0 & 33.5 & 25.8 & 30.3 & 13.5 & 16.4 & 22.6 & 30.0 & 22.4 & 25.4 & 24.9 & 31.2 & 13.6 & 16.0 \\
\textsc{Command R} \textsuperscript{\#11} & 11.1 & 15.2 & 18.6 & 26.1 & 13.0 & 17.1 & \phantom{0}5.2 & \phantom{0}7.4 & 10.4 & 17.0 & 10.3 & 12.3 & 14.9 & 20.2 & \phantom{0}7.3 & \phantom{0}9.4 \\
\midrule
\textsc{Qwen1.5-110b-chat} \textsuperscript{\#4} & 33.4 & 37.8 & \underline{36.2} & \underline{44.0} & 42.6 & 46.9 & 22.3 & 25.1 & 34.1 & 40.7 & \underline{34.8} & \underline{37.5} & 40.8 & 46.1 & 22.5 & 25.2 \\
\textsc{Qwen1.5-32b-chat} \textsuperscript{\#5} & 32.8 & 37.1 & 34.9 & 42.8 & \underline{43.2} & \underline{47.3} & 20.7 & 23.7 & 32.3 & 38.3 & 34.0 & 37.1 & 40.8 & 44.8 & 22.6 & 25.2 \\
\bottomrule
\end{tabular}
}
\vspace{-0.2cm}
\caption{Evaluation results on \lfrqa~test set. W and W+T indicate win and win+tie rate against \lfrqa's answers. LLM's answers are generated based on the top 5 passages. \textbf{bold} and \underline{underline} indicate the best and runner-up results. * means using the answer generation prompt w/o CoT. \textsuperscript{\#n} indicates the Elo ranking in Appendix Table~\ref{tab:elo}.}
\label{tab:eval-mbe-all}
\vspace{-0.3cm}
\end{table*}

\begin{table}[h]
\small
\setlength{\tabcolsep}{4pt}
\centering
\scalebox{0.9}{
\begin{tabular}{l|l|c|c|c|c}
\toprule
& \textbf{Answer} & & & Pearson & Cohen's \\
& \textbf{Pairs} & \textbf{Human} & \textbf{MBE} & Corr. & Kappa \\
\midrule
\midrule
& \rqa & \phantom{0}6.1 & \phantom{0}1.0 &  \\
(1) & \lfrqa & \textbf{83.9} & \textbf{95.6} & 0.82*** & 0.71 \\
& Tie & 10.0 & \phantom{0}3.4 & \\ 
\midrule
&\textsc{GPT-4} & \textbf{78.1} & \textbf{77.1} & \\
(2) &\rqa & 13.7 & 19.9 & 0.60*** & 0.54 \\
&Tie & \phantom{0}8.1 & \phantom{0}3.0 & \\ 
 \midrule
 \midrule
&\textsc{GPT-4} & 29.9 & 32.0 & \\
(3) &\lfrqa & \textbf{59.1} & \textbf{63.1} & 0.54*** & 0.44 \\
&Tie & 11.0 & \phantom{0}4.9 & \\ 
\midrule
\midrule
&\textsc{Mixtral} & 31.7 & 36.9 &\\
(4) &\lfrqa  & \textbf{54.4} & \textbf{59.7} & 0.54*** & 0.43  \\
&Tie & 13.9 & \phantom{0}4.4 &\\ 
\midrule
&\textsc{Llama-3} & 24.0 & 24.7 & \\
(5) &\lfrqa  & \textbf{65.0} & \textbf{71.3} & 0.52*** & 0.45 \\
&Tie & 11.0 & \phantom{0}4.0 &\\ 
\bottomrule
\end{tabular}
}
\vspace{-0.2cm}
\caption{Pairwise comparisons between \rqa, \lfrqa, and LLMs. \textsc{GPT-4}: \textsc{GPT-4-0125-preview}; \textsc{Mixtral}: \textsc{Mixtral-8x22b-Instruct}; \textsc{Llama-3}: \textsc{Llama-3-70b-Instruct}. Answer generated based on top 5 passages using ColBERT-v2. \textbf{MBE} stands for the model-based evaluator. All numbers are \% except for Pearson Corr. and Cohen's Kappa. *** indicates strong correlation with p-values $\ll 0.001$.}
\vspace{-0.7cm}
\label{tab:rqa-lfrqa-llm}
\end{table}

%% file: 06-results.tex
\section{Results and Analysis}
\label{sec:results}
Leveraging the evaluation framework described in Sec.~\ref{sec:eval}, we first show that \lfrqa's ground truth answers are dominantly preferred as answers than \rqa. Then, we use the same evaluation framework to establish a new leaderboard, \arena, aiming to reliably measure RAG-QA systems' performances across diverse domains. 

\begin{table*}[t]
\centering
\small
\scalebox{0.95}{
\setlength{\tabcolsep}{6pt}
\begin{tabular}{l|rrr|rrr|rrr}
\toprule

 & \multicolumn{3}{c|}{\textbf{(A)} \lfrqa~only} & \multicolumn{3}{c|}{\textbf{(B)} \lfrqa~ + } & \multicolumn{3}{c}{\textbf{(C)} \lfrqa~ + }  \\
  & \multicolumn{3}{c|}{pairs} & \multicolumn{3}{c|}{700 complete pairs} & \multicolumn{3}{c}{1400 complete pairs}  \\
  \midrule
 \textbf{\arena~Ranking} & \textbf{Rating} & \textbf{95\% CI} & \textbf{Votes} & \textbf{Rating} & \textbf{95\% CI} & \textbf{Votes} & \textbf{Rating} & \textbf{95\% CI} & \textbf{Votes} \\
 \midrule 
 \lfrqa & 1144 & +1/-1 & 176.7K & 1145 & +1/-1 & 176.7K & 1146 & +1/-1 & 176.7K \\
 \textsc{GPT-4o} & 1066 & +5/-5 & 16.1K & 1081 & +4/-4 & 23.1K & 1085 & +3/-3 & 30.1K \\
 \textsc{GPT-4-turbo} & 1050 & +5/-4 & 16.1K & 1058 & +4/-3 & 23.1K & 1065 & +3/-2 & 30.1K \\
 \textsc{Mixtral-8x22b} & 1049 & +4/-4 & 16.1K & 1059 & +3/-3 & 23.1K & 1063 & +3/-3 & 30.1K \\
 \textsc{Qwen1.5-110b-chat} & 1041 & +4/-4 & 16.1K & 1047 & +4/-3 & 23.1K & 1052 & +3/-3 & 30.1K \\
 \textsc{Qwen1.5-32b-chat} & 1036 & +6/-4 & 16.1K & 1034 & +4/-3 & 23.1K & 1037 & +3/-3 & 30.1K \\
 \textsc{GPT-4-0125-preview} & 1008 & +6/-5 & 16.1K & 1005 & +4/-4 & 23.1K & 1008 & +3/-3 & 30.1K \\
 \textsc{Mixtral-8x7b} & 991 & +4/-4 & 16.1K & 991 & +3/-4 & 23.1K & 987 & +3/-3 & 30.1K \\
 \textsc{Llama-3-70b} & 939 & +4/-5 & 16.1K & 931 & +4/-4 & 23.1K & 930 & +2/-3 & 30.1K \\
 \textsc{Command R+} & 938 & +5/-5 & 16.1K & 931 & +3/-4 & 23.1K & 924 & +3/-3 & 30.1K \\
 \textsc{Llama-3-8b} & 924 & +6/-6 & 16.1K & 910 & +4/-4 & 23.1K & 903 & +4/-3 & 30.1K \\
 \textsc{Command R} & 816 & +8/-6 & 16.1K & 802 & +5/-5 & 23.1K & 796 & +4/-4 & 30.1K \\
 \bottomrule
 \end{tabular}
 }
 \caption{Elo rating \textbf{including} ``couldn't find answer'' responses. ``\lfrqa~only'' indicates the pairs that always include an \lfrqa~answer. ``\lfrqa~ + N complete pairs'' means we subsample N additional pairs evenly across 7 domains and conduct comparison for all pairs of models. We have 11 models, so the total new pairs are ${\frac{N\times11\times10}{2}}$.}
\label{tab:elo}
 \end{table*}

\subsection{LFRQA v.s. RobustQA}
\label{sec:lfrqa-rqa}

In Sec.~\ref{sec:analysis}, we demonstrate \lfrqa's advantages via data statistics. Here, we show a more rigorous study to highlight the benefits. We subsample 700 queries (100 from each of the 7 domains) and conduct pairwise preference comparisons using both human and model-based evaluations. We compare three types of answers: 1) \rqa: concatenation of its extractive answers, separated by "\textbackslash n"; 2) \lfrqa: long-form answers in this work; 3) \textsc{GPT-4}'s answers based on the top 5 retrieved passages.

Table~\ref{tab:rqa-lfrqa-llm} shows that when compared directly in Row (1), \lfrqa~dominates \rqa. When comparing with \textsc{GPT-4} in Row (2)-(3),  \lfrqa~significantly out-performs \textsc{GPT-4}, but \rqa~significantly under-performs. These results show strong evidence that \lfrqa's answers can serve as better ground-truth than \rqa.

\subsection{Quality of Model-based Evaluator}
To build \arena~on the \lfrqa~test set, we need a scalable evaluation method to benchmark various LLMs. We rely on model-based evaluation to achieve this goal. Before showing the final dashboard results, we check the quality of our selected evaluator (\textsc{GPT-4-0125-preview}) in Table~\ref{tab:rqa-lfrqa-llm}. 

To alleviate model bias, we use three LLMs' answers as benchmark data, and the query set is the same 700 subsample above. Row (3)-(5) use \textsc{GPT-4-0125-preview}, \textsc{Mixtral-8x22b-Instruct} and \textsc{Llama-3-70b-Instruct}, respectively. All answers are generated based on the top 5 passages. 

We observe that LLM evaluators' numbers align well with the average human scores, except that LLMs tend to predict less ``tie.'' Most importantly, all Pearson Correlation \cite{freedman2007statistics} are above 0.52 (with p-values $\ll 0.01$), and all Cohen's Kappa \cite{cohenKappa} are above 0.43, both showing strong agreement between model and human judgments. In Appendix Table~\ref{tab:eval-model-select}, we show correlation numbers using alternative LLMs as evaluators, but none of them works better than a single \textsc{GPT-4-0125-preview} model, which we select as our best quality evaluator for \arena.

\subsection{RAG-QA Arena}
\label{sec:res:rag-qa}
Finally, we show \arena's benchmark results. In Table~\ref{tab:eval-mbe-all} we report each model's win and win+tie rate against \lfrqa. 

\mypar{Dashboard leaders.} \textsc{GPT-4o} leads the dashboard, with \textsc{GPT-4-turbo} and \textsc{Mixtral-8x22b-Instruct} as close runners-up. \textsc{GPT-4o} performs the best for [BI], [LI], [RE] and [WR] domains, \textsc{Mixtral-8x22b-Instruct} leads in [FI] and [SC], and \textsc{GPT-4-turbo} champions in [TE].

\mypar{Impact of ``no answer found.''} In RAG-QA, we rely on a passage retriever to provide context, which could be irrelevant. Our prompt (Appendix Table~\ref{tab:answer-gen}) asks an LLM to refrain from answering if it ``couldn't find an answer.'' When we use this answer generation prompt with CoT (the last two lines in the prompt), \textsc{GPT-4o} produces 48.3\% ``I couldn't find an answer'' responses (Appendix Table~\ref{tab:no-answer}). We randomly sample 20 such examples, and surprisingly found that in 16 cases, \textsc{GPT-4o} puts an answer in its \textbf{<thinking>} process, but continues to generate ``I couldn't find an answer.'' Fig.~\ref{fig:errors-1}-\ref{fig:errors-2} show four such examples in comparison with other LLMs' answers with the same prompt, and \textsc{GPT-4o}'s new answers without CoT. As the answer generation prompt with CoT only fails for \textsc{GPT-4o}, we remove CoT for \textsc{GPT-4o}, which improves its answer format and reduces the ``no-answer'' ratio to the level similar to other competitive models.

These results raise a research question about the impact of prompt engineering. We emphasize that the goal of \arena~is to propose a reliable evaluation framework, not to conduct extensive prompt engineering or model training to pursue the best RAG-QA system. We provide a dev set of \lfrqa, which can be leveraged in future research for model development purposes.

\mypar{Elo rating.} Table~\ref{tab:eval-mbe-all} shows dashboard results of win and win+tie ratio against \lfrqa. We can further convert these pairwise comparisons into Elo ratings similar to Chatbot Arena \cite{Chiang2024ChatbotAA}. Table~\ref{tab:elo} reports our Elo ranking. The leftmost column shows \arena's ranking based on win ratio: \lfrqa~is ranked on the top followed by \textsc{GPT-4o}, \textsc{GPT-4-turbo}, etc. Column (A) uses the same data in Table~\ref{tab:eval-mbe-all}, but here we only have comparisons between \lfrqa~and other LLMs' responses, i.e. there are no direct comparisons between LLMs' responses. Thus, the total number of votes for are 176.7K for \lfrqa, and 16.1K for other LLMs. The new ranking based on Elo rating aligns with \arena, but the 95\% confidence interval (CI) is not yet able to separate all different model pairs. 

In Column (B), we add pairwise comparisons for all unique model pairs on 700 randomly sampled queries across seven domains (100 for each domain). This increases the votes for all LLMs compared to 23.1K. We continue to add 700 more queries in Column (C), which further increases the votes for all LLMs to 30.1K. We rank these pairs with the same LLM evaluator as in the main result table, and found that the ranking in (C) based on Elo rating aligns perfectly with \arena, and as we increase the pairs, the 95\% CI can finally separate different models.

In general, the total added preference pairs are $\frac{N\times K \times (K-1)}{2}$, where $K$ is the number of models in Table~\ref{tab:eval-mbe-all}, and $N$=700 and 1400 for Column (B) and (C), respectively. With 43.6\% increase of total pairs (and thus the compute), the final ranking is identical with \arena~based on win ratio, and only differs only slightly with the win+tie ranking in Table~\ref{tab:eval-mbe-all} by flipping the order of \textsc{llama-3-70b} and \textsc{Command R+}. These results present additional evidence that our approach of using \lfrqa~only for pairwise comparisons is reliable. Furthermore, it reduces the computational costs from $\mathcal{O}(K^2)$ to  $\mathcal{O}(K)$ as we now only need to compare each LLM response once with the ground-truth in \lfrqa.

\mypar{Impact of the number of passages.} In Appendix Table~\ref{tab:eval-mbe-10psg}, we compare the top 3 LLMs by increasing the number of retrieved passages from 5 to 10. Doubling the number of passages (with extra costs) increases RAG-QA performances significantly. We also find that both \textsc{GPT-4} models' improvements are greater than \textsc{Mixtral-8x22b-Instruct}, showing their superior capability to understand long context and identify useful information from noise. The best win rate of \textsc{GPT-4o} against \lfrqa~is 41.3\%, which is 13.7\% points lower than \lfrqa~answers' win rate against \textsc{GPT-4o}. This result shows that \lfrqa's answer quality is difficult to surpass, further justifying using it as an evaluation target.

\mypar{Impact of model sizes.} For the non-GPT LLM family, more parameters lead to better performances, but a larger increase in model sizes does not always indicate greater performance gains in our study. For example, the two \textsc{Qwen1.5} models have the second-largest difference, but the lift from the smaller to the larger model is marginal. We leave more rigorous investigations to future research.

%% file: 07-discuss.tex
\section{Discussion}
\subsection{LLM as Annotators}

\begin{table}[t]
\small
\centering
\scalebox{0.95}{
\begin{tabular}{l|c|c|c}
\toprule
     & & \textbf{Citation} & \textbf{} \\
 & \textbf{Completeness} & \textbf{Accuracy} & \textbf{Helpfulness} \\
\midrule
\lfrqa & 90.8 & 88.9 & 48.1 \\
\textsc{GPT-4} & 75.3 & 65.5 & 35.2 \\
\bottomrule
\end{tabular}
}
\caption{Comparison between human and GPT-4 annotations. All numbers are in \%.}
\label{tab:human-gpt4}
\vspace{-0.3cm}
\end{table}

Using large language models to provide annotations has been explored in previous works \cite{llmAnnotate}. It could provide a more scalable solution than human annotations but can suffer from hallucination and accuracy issues that require human validations \cite{Huang2023ASO}. We also experimented with LLM as annotators before we start human annotations. We subsample 100 queries from \lfrqa~and prompt \textsc{GPT-4-0125-preview} to follow the similar procedure in Sec.~\ref{sec:annotation} to combine answers (Appendix Table~\ref{tab:answer-combine}). Then we request our data linguists to compare \lfrqa~and \textsc{GPT-4} annotations based on 1) Completeness: whether all \rqa~answers are integrated into the final answers; 2) Citation Accuracy: whether citations in answers pointing to the right documents; 3) Helpfulness: defined the same as in Sec.~\ref{sec:eval}. Table~\ref{tab:human-gpt4} shows \lfrqa~out-performs \textsc{GPT-4} annotations by 15.5\%, 23.4\% and 12.9\% for the three dimensions, respectively, suggesting human annotations are both valuable and necessary for our task.

\subsection{Alternative Evaluation Approaches}
\mypar{Using retrieved passages.} \arena~leverages only \lfrqa's annotations as ground-truth to directly evaluate LLM responses, and we explain this design choice in Sec.~\ref{sec:eval} that \lfrqa~consists of complete and coherent answers that can be viewed as high-quality summary of all available answers in the entire corpus. This enables us to not show retrieved passages as they 1) increase the input length and thus the latency of an evaluator; and 2) they could contain incorrect information due to retrieval error, which mislead evaluators.

\mypar{Using \lfrqa~as references.} We can also use \lfrqa's annotations as references when constructing the prompt for pairwise evaluation. That is, we can potentially compare a pair of LLMs' responses by comparing them both against the references in a single trial. However, this approach would still require the similar $\mathcal{O}(K^2)$ pairs as in the Elo rating, which is not as efficient as our proposed \arena~framework.

For these reasons, we do not adopt the above two evaluation approaches. It is conceivable that prompt engineering, in-context example selections and even task specific evaluator training could further enhance alignments with human judges. We leave them for future research efforts.

%% file: 08-related.tex
\section{Related Work}
\label{sec:related}

\mypar{RAG-QA} has been widely studied. Prior datasets are limited in the evaluation as their corpus relies heavily on Wikipedia and the answers are mostly short and extractive \cite{rajpurkar-etal-2016-squad, kwiatkowski-etal-2019-natural, qampari}. \rqa~and \textsc{MultiHop-RAG} \cite{MultiHop-RAG} address the single domain issue, but still adopt short, extractive answers, which is not as suitable as \lfrqa~to evaluate modern LLMs that generate long-form answers.

\mypar{Longform QA datasets} have been proposed in prior work. ELI5 \cite{fan-etal-2019-eli5} and \textsc{LongFact} \cite{LongFact} contain answers that are either not annotated directly on the corpus, and or not created by humans. \citet{krishna-etal-2021-hurdles} also points out that \textsc{ELI5}'s small validation set has significant leakage from its train set. \textsc{ASQA} \cite{stelmakh-etal-2022-asqa} is the most similar data to our work, but its corpus is in the single Wikipedia domain. \lfrqa~is by far the RAG-QA dataset with the most comprehensive long-form answers.

\mypar{Pairwise preference} is now a standard way to evaluate LLMs. It allows direct comparison between two responses \cite{Chiang2024ChatbotAA, wildbench2024}. \arena~is unique by always including a high-quality human annotated \textbf{LFRQA}~answer, thereby making the evaluation more trustworthy.

%% file: 09-conclude.tex
\section{Conclusion}
We create \lfrqa, the first multi-domain dataset with coherent long-form answers to reliably benchmark RAG-QA. We propose a reliable LLM-based evaluation framework, \arena, that enables direct comparisons between LLMs' answers and \lfrqa, which we believe will facilitate the evaluation RAG-QA in the era of LLMs.

%% file: 99-appendix.tex
\section{Appendix}
\label{sec:appendix}


\begin{table*}[!t]
\centering
\small
\scalebox{0.85}{
\begin{tabular}{l|l|l|rrr|rr|rr}
\toprule
 & & & \multicolumn{3}{c|}{\textbf{Dev Set}} & \multicolumn{2}{c}{\rqa} & \multicolumn{2}{|c}{\lfrqa} \\
\midrule
 \textbf{Domain} & \textbf{Source} & \textbf{Label} & \multicolumn{1}{c}{\textbf{$|Q|$}}  & \multicolumn{1}{c}{\textbf{$|D|$}} & \multicolumn{1}{c}{\textbf{$|P|$}}  & \multicolumn{1}{|c}{$A/Q$} & \multicolumn{1}{c}{\textbf{$W/A$}} & \multicolumn{1}{|c}{\textbf{$A/Q$}} & \multicolumn{1}{c}{\textbf{$W/A$}}  \\
 \midrule
Lifestyle & LoTTE & [LI] & \multicolumn{1}{r}{2,151} & 268,893 &  \multicolumn{1}{r|}{597,729} & 5.9 & 5.9 & 1.0 & 102.2 \\
Recreation & LoTTE & [RE] & \multicolumn{1}{r}{2,325} & 263,025 &  \multicolumn{1}{r|}{731,124} & 6.3 & 7.2 & 1.0 & 112.8 \\
Technology & LoTTE & [TE] & \multicolumn{1}{r}{2,223} & 1,000,000 & \multicolumn{1}{r|}{1,707,346} & 5.4 & 8.9 & 1.0 & 83.3\\
Science & LoTTE & [SC] & \multicolumn{1}{r}{2,137} & 343,642 & \multicolumn{1}{r|}{854,756} & 4.8 & 6.4 & 1.0 & 100.6\\
Writing & LoTTE & [WR] & \multicolumn{1}{r}{1,972} & 277,072 & \multicolumn{1}{r|}{713,692} & 7.0 & 7.7 & 1.0 & 109.6\\
\bottomrule
\end{tabular}
}
\caption{Data summary for the dev set. Based on \rqa, there is no dev split for BioASQ and FiQA data.}
\label{tab:data-stats-dev}
\end{table*}


\subsection{Dev Set of LFRQA}
\label{sec:dev-data}
We do not use dev set to fit the scope of the paper, but we provide an additional 10K queries for future model developments, including prompting engineering and training. Details are shown in Table~\ref{tab:data-stats-dev}. All data are collected using the same process and quality control described in Sec.~\ref{sec:data}.

\subsection{Data License}
\label{sec:license}
\lfrqa~is created based on the following datasets. We make sure to follow the data distribution license for our usage of the data.
\begin{itemize}
    \item \textbf{FiQA:} no license provided, but all data information can be found one the official website\footnote{https://sites.google.com/view/fiqa/home}.
    \item \textbf{LoTTE:} MIT license\footnote{https://github.com/stanford-futuredata/ColBERT}.
    \item \textbf{BioQAS:} CC BY 2.5 license\footnote{http://participants-area.bioasq.org/datasets/}.
    \item \textbf{\textsc{RobustQA}:} Apache-2.0 license\footnote{https://github.com/awslabs/robustqa-acl23}.
\end{itemize}


\subsection{Annotation Interface}
\label{sec:interface}
\begin{figure*}[t]
    \centering
\includegraphics[trim={0cm 0cm 0cm 0cm}, clip, angle=-90, width=\textwidth]{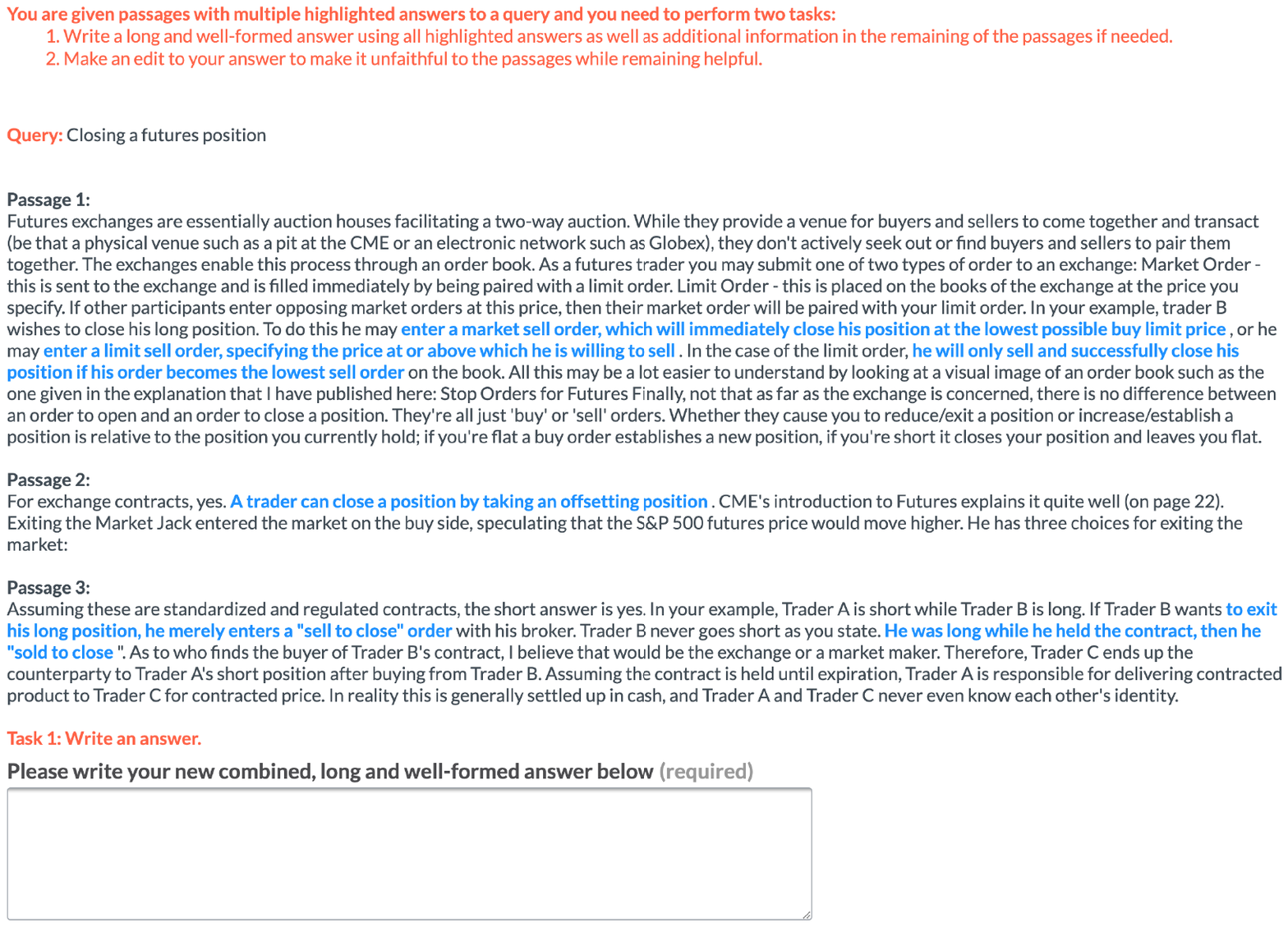}
\caption{Annotation Interface}
\label{fig:anno-ui}
\end{figure*}

Our annotation interface can be found in Fig.~\ref{fig:anno-ui}. Annotators use this UI to write long-form answers. Blue highlights are the original answers from \rqa.


\subsection{Answer Generation Prompts}
\label{sec:ans-gen}
Table~\ref{tab:answer-gen} shows our answer generation prompt. The last two lines ``First, think step-by-step...'' are what we refer to as the CoT prompt, which we remove for \textsc{GPT-4o}.


\subsection{Evaluation Interface}
\label{sec:eval-interface}
\begin{figure*}[t]
    \centering
\includegraphics[trim={3cm 0cm 3cm 0cm}, clip, angle=-90, width=0.99\textwidth]{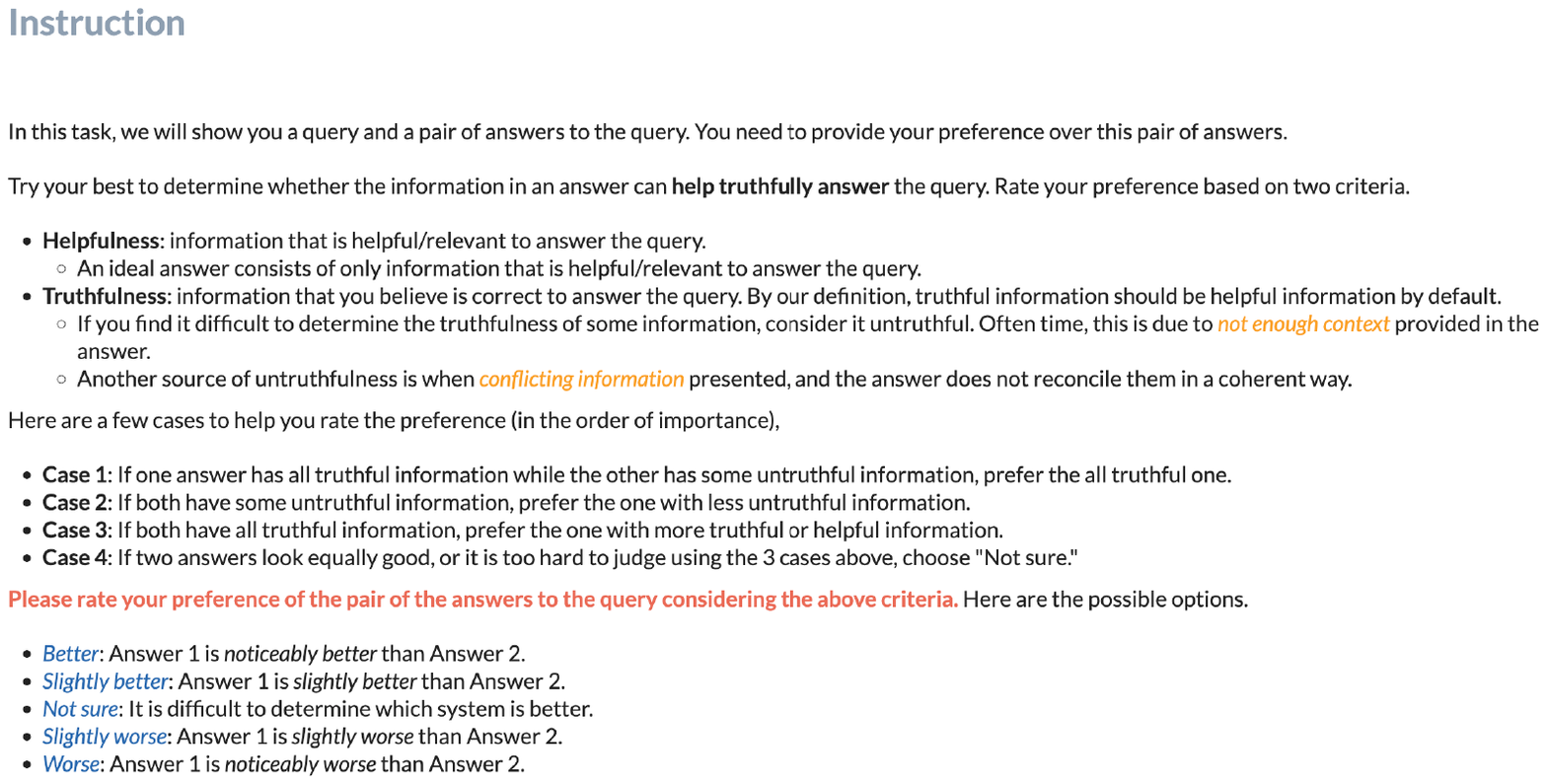}
\includegraphics[trim={2cm 0cm 2cm 0cm}, clip, angle=-90, width=0.99\textwidth]{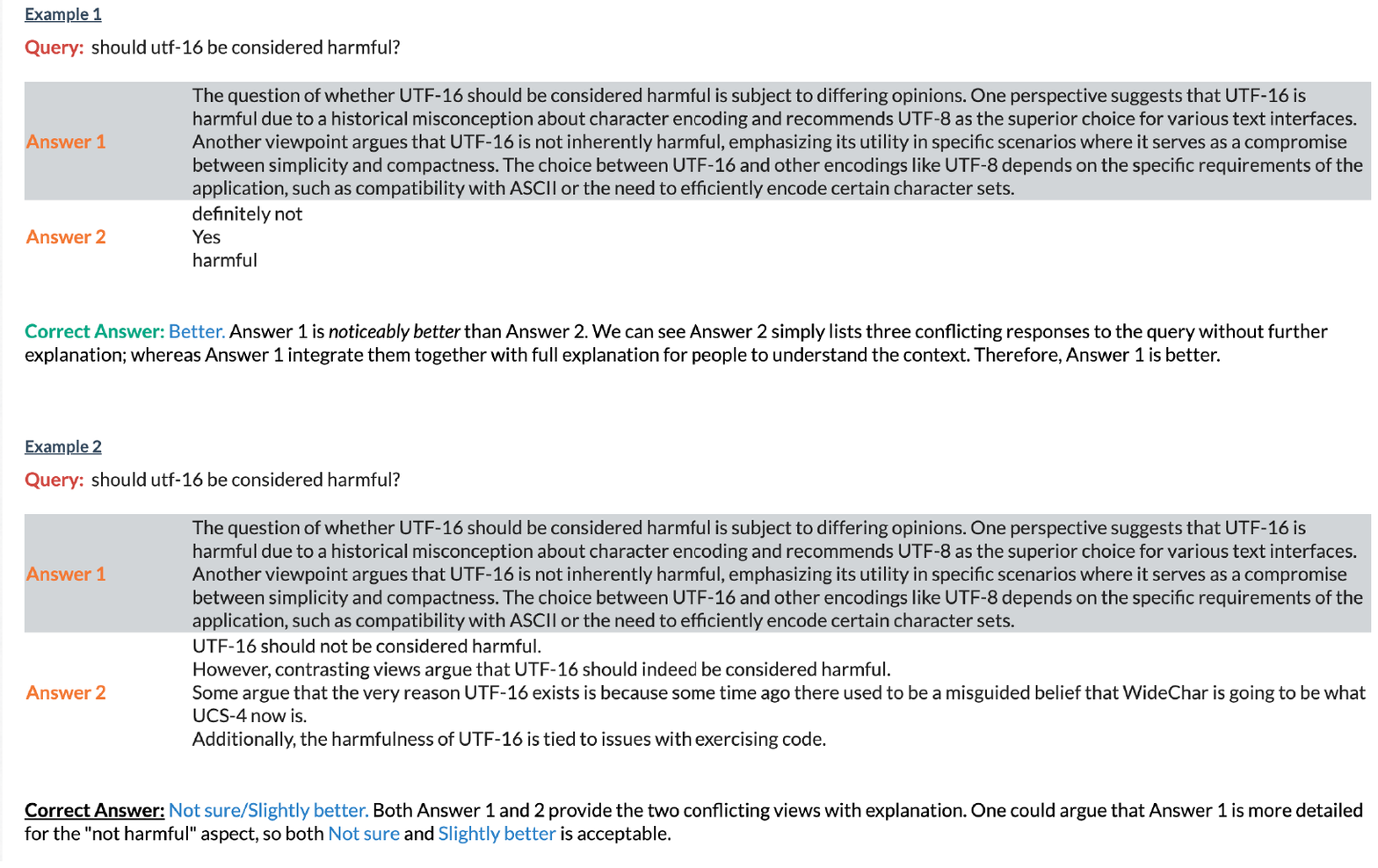}
\caption{Pairwise preference: instruction.}
\label{fig:eval-ui-a}
\end{figure*}

\begin{figure*}[t]
    \centering
\includegraphics[trim={2cm 0cm 2cm 0cm}, clip, angle=-90, width=0.99\textwidth]{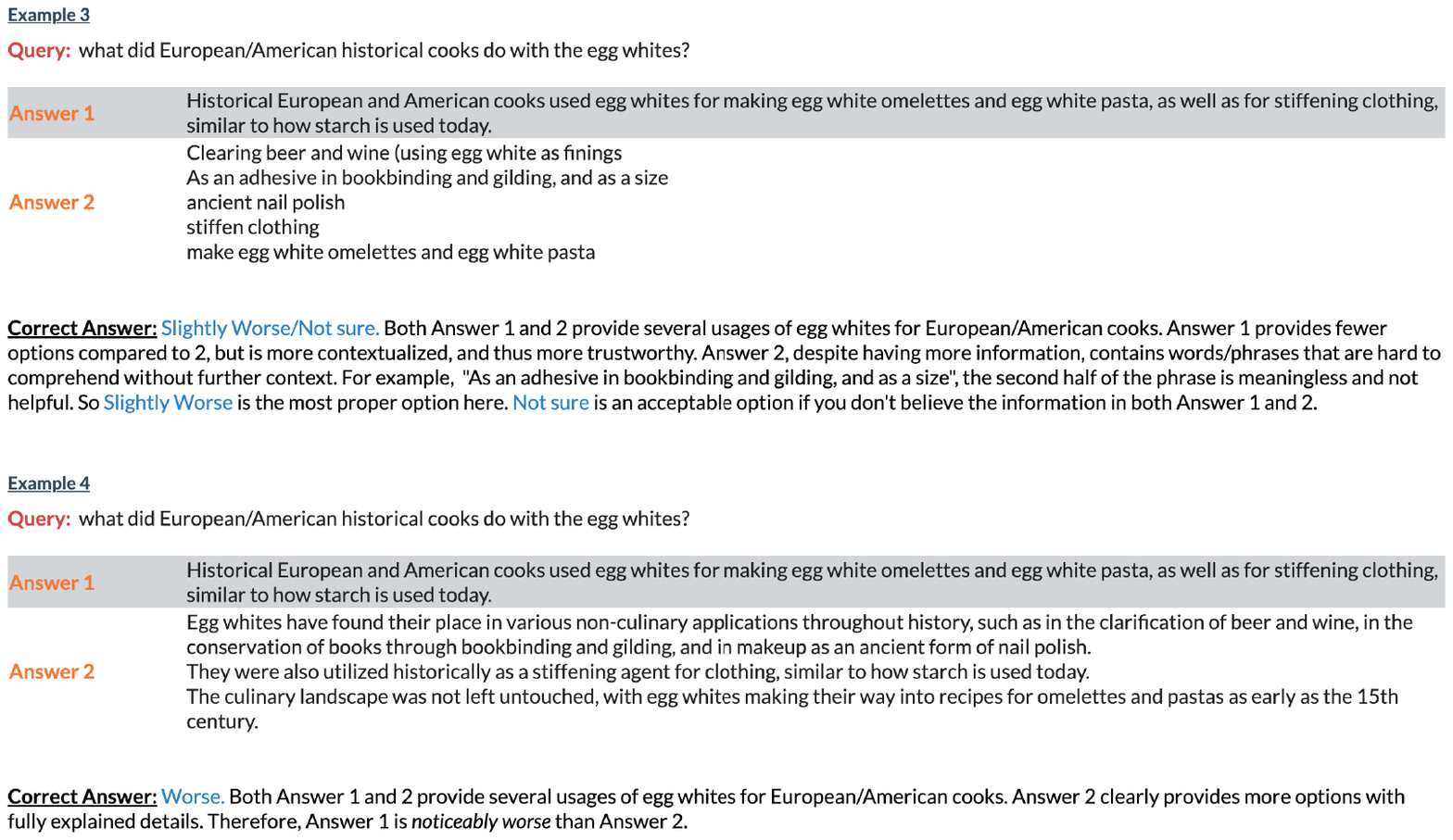}
\includegraphics[trim={4cm 0cm 4cm 0cm}, clip, angle=-90, width=0.99\textwidth]{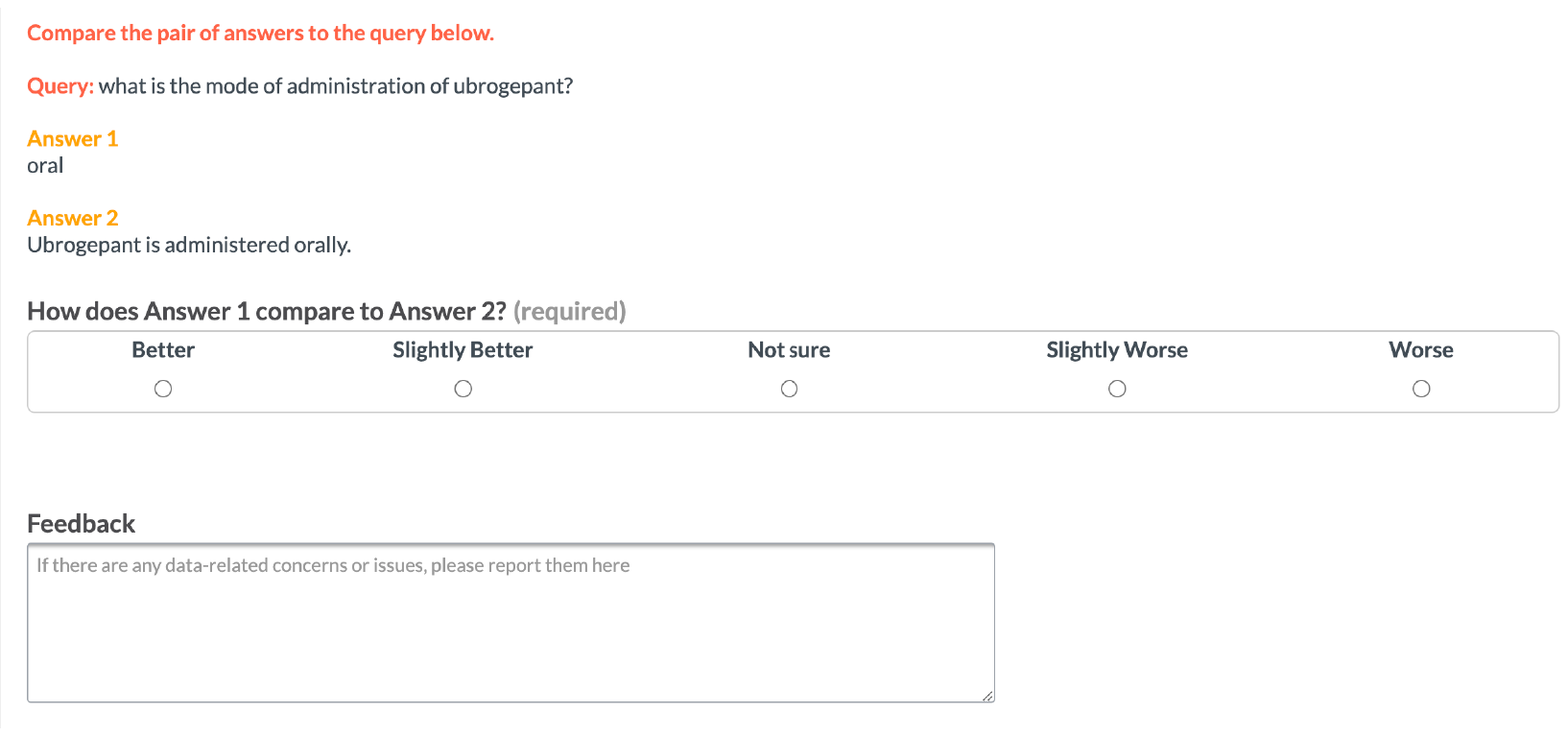}
\caption{Pairwise preference: annotation interface.}
\label{fig:eval-ui-b}
\end{figure*}

We collect human pairwise preference data to benchmark our LLM-based evaluators. Fig.~\ref{fig:eval-ui-a}-\ref{fig:eval-ui-b} show the details.

\subsection{Human Evaluation Instructions}
\label{sec:metric-define}
\mypar{Helpfulness:} Information that is helpful/relevant to answer the query. An ideal answer consists of only information that is helpful/relevant to answer the query \cite{llama-2, Bai2022TrainingAH}.

\mypar{Truthfulness:} Information that is correct to answer the query. By our definition, truthful information should also be helpful information. If it is difficult to determine the truthfulness of some information, we consider it untruthful. Sometimes, this is due to not enough context provided in the answer. Another source of untruthfulness is when conflicting information is presented, and the answer does not coherently reconcile them \cite{truthfulqa, truthfuleval}.

\mypar{Completeness:} include as many helpful and truthful information.

Here are the details of our instructions.

\mypar{1.} If one answer has all truthful information while the other has some untruthful information, prefer the all-truthful one. \mypar{2.} If both have some untruthful information, prefer the one with less untruthful information. \mypar{3.} If both have all truthful information, prefer the one with more truthful or helpful information. \mypar{4.} If two answers look equally good, or it is too hard to differentiate, choose ``Not sure.''

As the annotation UI shows, the actual ratings are ``Better,'' ``Slightly Better,'' ``Tie,'' ``Slightly Worse'' and ``Worse''. We merge ``Better'' and ``Slightly Better,'' and  ``Slightly Worse'' and ``Worse'' when computing correlation with model-based evaluators.

\subsection{Evaluation Model Selection}
Table~\ref{tab:eval-model-select} shows Pearson Corr. between human and LLM-based evaluators. We find using \textsc{GPT-4-0125-preview} alone achieves the best outcome, and thus, use it for \arena~evaluations.

\begin{table}[h]
\footnotesize
\centering
\scalebox{0.8}{
\begin{tabular}{l|c|c|c}
\toprule
& \lfrqa~v.s. & \lfrqa~v.s. & \lfrqa~v.s. \\
\textbf{Evaluator Models} & \textsc{GPT-4} & \textsc{Mixtral} & \textsc{Llama-3}\\
\midrule
Llama-3-70b & 0.52 & 0.51 & 0.48 \\
Mixtral-8x22b & 0.53 & 0.49 & 0.50 \\
GPT-4-turbo & 0.47 & 0.49 & 0.48 \\
GPT-4o & 0.43 & 0.46 & 0.45 \\
GPT-4-0125-preview & \textbf{0.54} & \textbf{0.54} & \textbf{0.52} \\ 
\midrule
Ensemble & 0.53 & 0.53 & \textbf{0.52} \\
\bottomrule
\end{tabular}
}
\caption{Correlation between human judges and leading LLMs as evaluators. Column titles are answer pairs. \textsc{GPT-4}: GPT-4-0125-preview; \textsc{Mixtral}: Mixtral-8x22b-Instruct-v0.1; \textsc{Llama-3}: Llama-3-70b-Instruct. The ensemble takes the majority vote out of the best three models. }
\vspace{-0.5cm}
\label{tab:eval-model-select}
\end{table}


\subsection{No Answer Ratio}
Table~\ref{tab:no-answer} shows the ratio of ``I couldn't find an answer" in each LLM's answers. 


\subsection{Pairwise Evaluation Prompts}
\label{sec:pair-eval}

Table~\ref{tab:pair-eval-instruction}-\ref{tab:pair-eval-examples} show the details of the pairwise evaluation prompts including instruction, example prompt template, and in-context-learning examples. 


\subsection{Error Analysis}
\begin{figure*}
    \centering
\includegraphics[trim={0.3cm 0cm 0.3cm 0cm}, clip, angle=-90, width=0.99\textwidth]{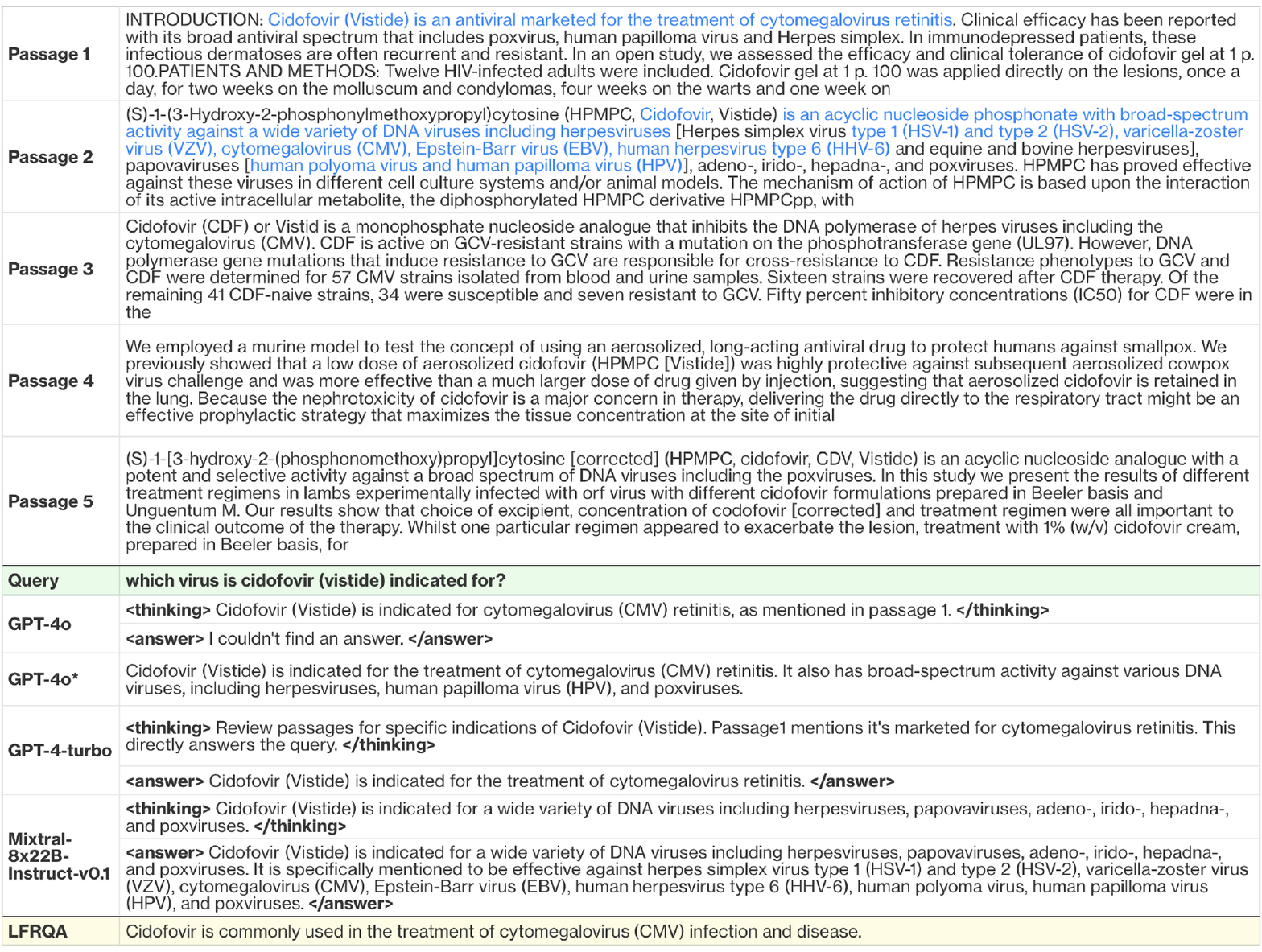}
\includegraphics[trim={1cm 0cm 1cm 0cm}, clip, angle=-90, width=0.99\textwidth]{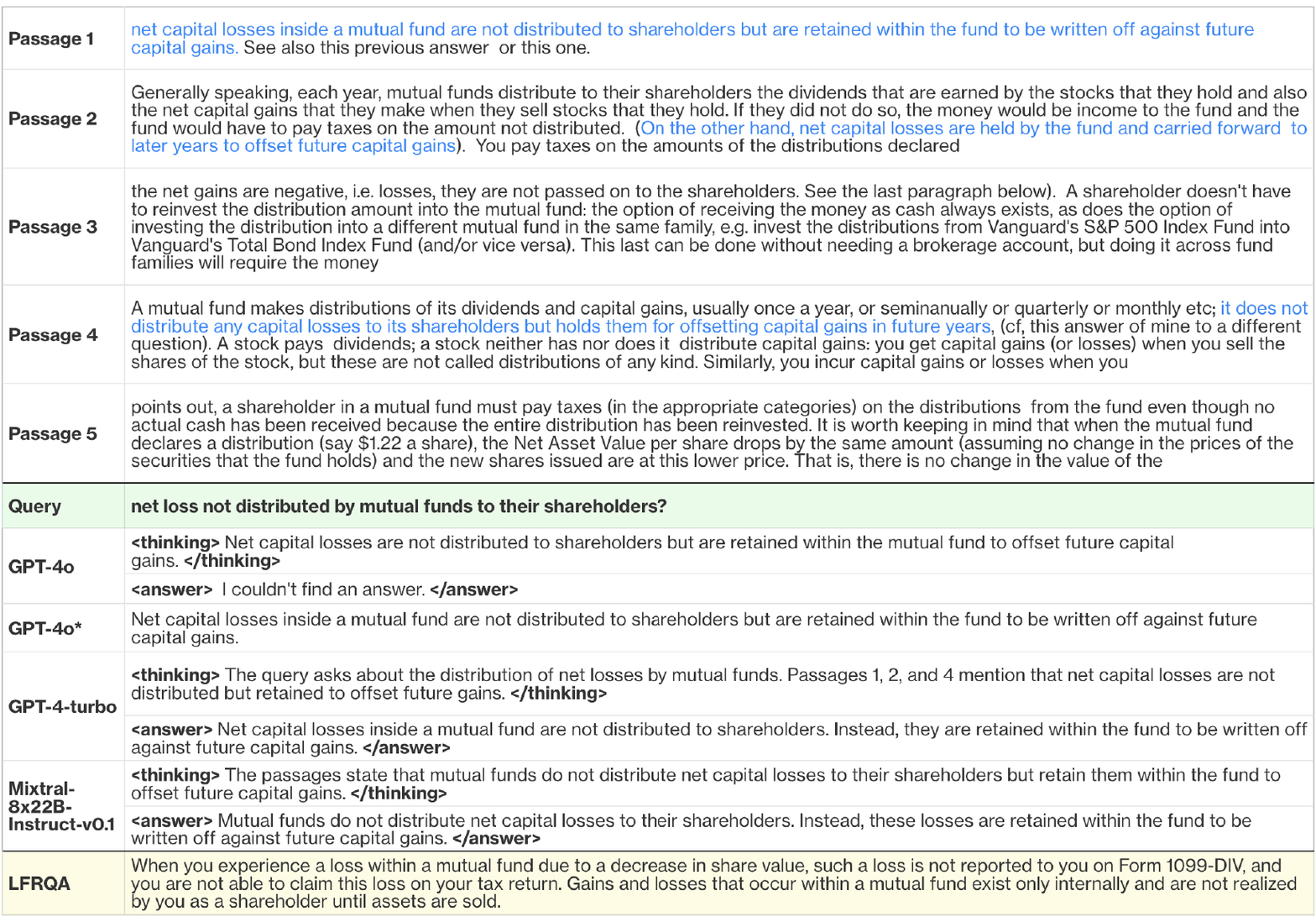}
\caption{\textsc{GPT-4o}'s original ``I couldn't find an answer'' response compared with \textsc{GPT-4-Turbo} and \textsc{Mixtral-8x22b}, and its answer without CoT (*). Blue highlights in the passages indicate helpful information to answer the query. \textbf{<answer>} tags are added to help differentiate from \textbf{<thinking>}.}
\label{fig:errors-1}
\end{figure*}

\begin{figure*}
    \centering
\includegraphics[trim={0.7cm 0cm 0.7cm 0cm}, clip, angle=-90, width=0.99\textwidth]{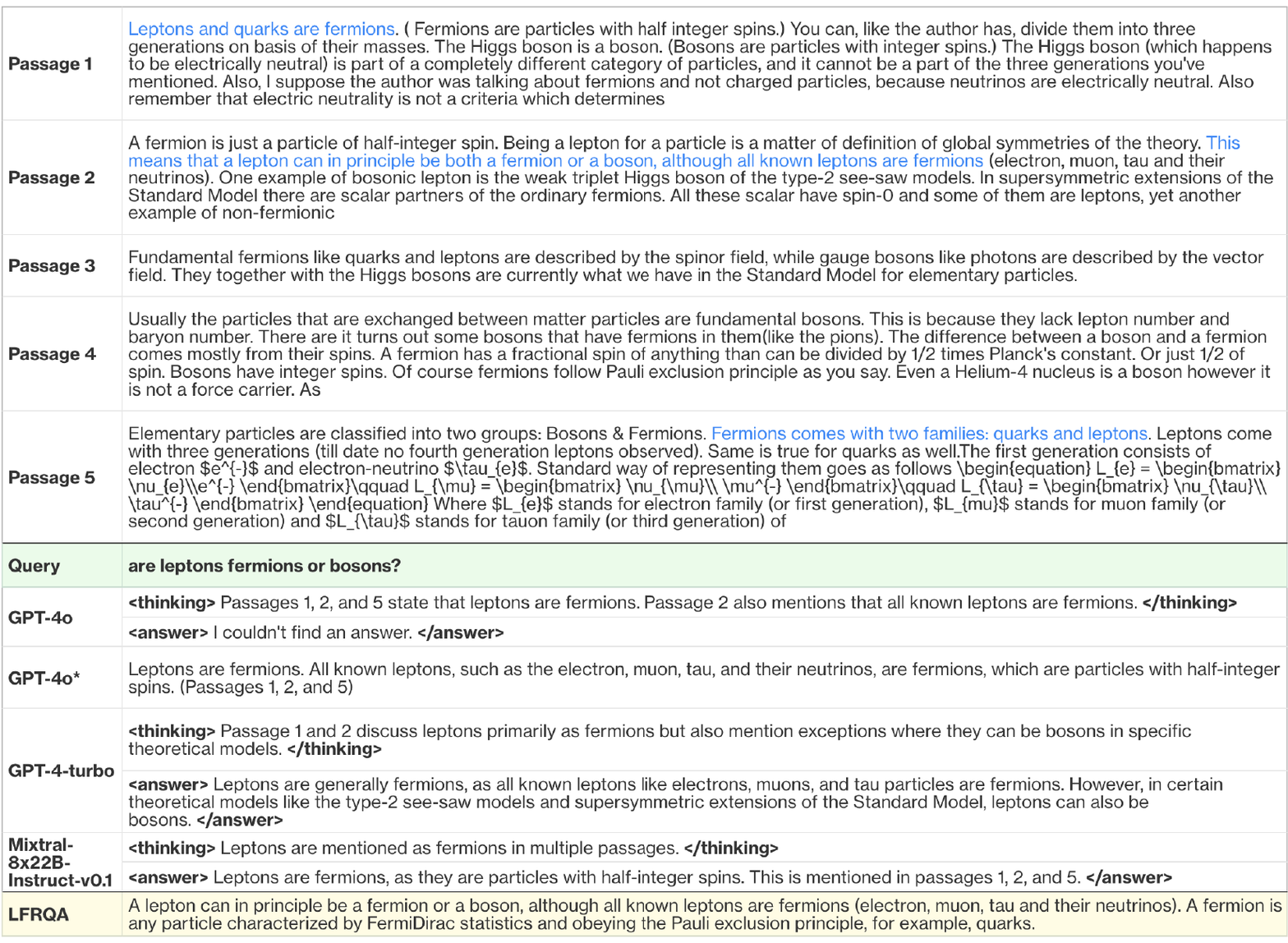}
\includegraphics[trim={0.7cm 0cm 0.7cm 0cm}, clip, angle=-90, width=0.99\textwidth]{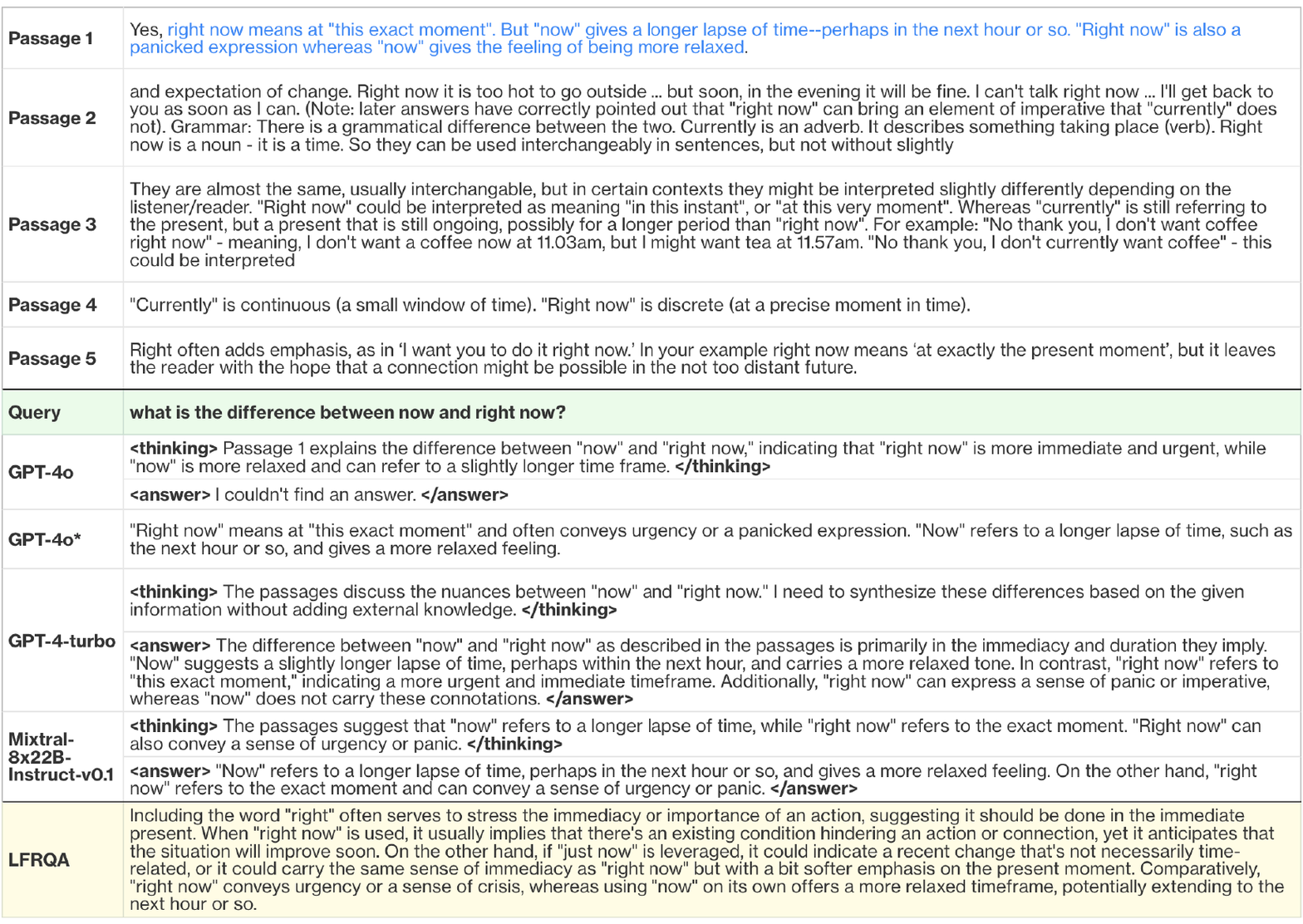}
\caption{\textsc{GPT-4o}'s original ``I couldn't find an answer'' response compared with \textsc{GPT-4-Turbo} and \textsc{Mixtral-8x22b}, and its answer without CoT (*). Blue highlights in the passages indicate helpful information to answer the query. \textbf{<answer>} tags are added to help differentiate from \textbf{<thinking>}.}
\label{fig:errors-2}
\end{figure*}

Fig.~\ref{fig:errors-1}-\ref{fig:errors-2} show examples where the original \textsc{GPT-4o} puts an answer in its \textbf{<thinking>}, but generate ``I couldn't find an answer." It is fixed by removing CoT prompt. We also compare them with \textsc{Mixtral-8x22b}, \textsc{GPT-4-turbo} and \lfrqa's answers.

\begin{table*}[t]
\centering
\small
\scalebox{0.85}{
\begin{tabular}{l|c|c|c|c|c|c|c|c}
\toprule

 & \textbf{Overall} & [BI] & [FI] & [LI] & [RE] & [TE] & [SC] & [WR] \\
\midrule
\textsc{GPT-4-turbo} & 14.1 & 10.9 & 19.8 & 12.2 & 15.9 & 11.2 & 12.3 & 12.1 \\
\textsc{GPT-4o} & 48.3 & 32.4 & 55.2 & 50.3 & 52.8 & 43.4 & 49.4 & 48.4 \\
\textsc{GPT-4o} * & 16.9 & 10.4 & 26.8 & 13.9 & 19.8 & 11.2 & 14.7 & 13.9  \\
\textsc{GPT-4-0125-preview} & 15.8 & 11.0 & 21.9 & 13.5 & 19.1 & 12.6 & 13.5 & 14.1 \\
\midrule
\textsc{Mixtral-8x22b} & 9.9 & 10.1 & 13.6 & 9.1 & 13.7 & 7.1 & 6.6 & 6.0 \\
\textsc{Mixtral-8x7b} & 18.5 & 15.4 & 24.6 & 17.8 & 23.7 & 13.3 & 15.3 & 15.0 \\
\midrule
\textsc{Llama-3-70b} & 25.7 & 17.7 & 35.9 & 24.4 & 29.2 & 20.5 & 23.3 & 21.2 \\
\textsc{Llama-3-8b} & 12.6 & 10.0 & 15.9 & 11.5 & 16.2 & 8.4 & 12.9 & 11.0 \\
\midrule
\textsc{Command R+} & 14.4 & 10.8 & 21.7 & 12.5 & 16.1 & 10.1 & 14.7 & 10.5 \\
\textsc{Command R} & 6.5 & 5.0 & 10.8 & 6.1 & 6.8 & 3.4 & 4.6 & 5.0  \\
\midrule
\textsc{Qwen1.5-110b-chat} & 11.5 & 10.8 & 15.9 & 10.7 & 13.8 & 6.8 & 9.0 & 10.0  \\
\textsc{Qwen1.5-32b-chat} & 8.9 & 7.4 & 13.6 & 7.5 & 11.8 & 5.0 & 7.0 & 6.5 \\
\bottomrule
\end{tabular}
}
\caption{No answer ratio for the entire \lfrqa's test set. * means using answer generation prompt without CoT.}
\label{tab:no-answer}
\end{table*}

\begin{table*}[t]
\centering
\small
\scalebox{0.75}{
\setlength{\tabcolsep}{4.5pt}
\begin{tabular}{l|cc|cc|cc|cc|cc|cc|cc|cc}
\toprule

 & \multicolumn{2}{c|}{\textbf{Overall}} & \multicolumn{2}{c|}{[BI]} & \multicolumn{2}{c|}{[FI]} & \multicolumn{2}{c|}{[LI]} & \multicolumn{2}{c|}{[RE]} & \multicolumn{2}{c|}{[TE]} & \multicolumn{2}{c|}{[SC]} & \multicolumn{2}{c}{[WR]} \\
\midrule
 & W & W+T & W & W+T & W & W+T & W & W+T & W & W+T & W & W+T & W & W+T & W & W+T \\
\midrule
\textsc{GPT-4o}* w/ \textbf{5 psgs}  & 36.9 & 41.0 & 52.9 & 59.3 & 38.4 & 42.3 & 25.1 & 27.9 & 40.4 & 46.4 & 35.6 & 38.8 & 42.8 & 47.6 & 28.4 & 31.1 \\

\textsc{GPT-4o}* w/ \textbf{10 psgs}  & \textbf{41.3} & \underline{45.0} & \textbf{59.1} & \textbf{64.4} & 45.7 & 49.4 & \underline{27.4} & \underline{30.0} & \textbf{43.7} & \textbf{49.9} & \underline{39.6} & \underline{42.2} & \underline{46.0} & \underline{50.0} & \underline{30.7} & \underline{33.1} \\

\midrule

\textsc{GPT-4-turbo} w/ \textbf{5 psgs} & 34.4 & 39.1 & 36.0 & 43.9 & 40.6 & 45.1 & 23.2 & 26.1 & 36.7 & 44.1 & 36.6 & 39.6 & 42.6 & 47.9 & 26.2 & 29.6 \\

\textsc{GPT-4-turbo} w/ \textbf{10 psgs} & \underline{40.6} & \textbf{45.4} & 40.8 & 49.5 & \textbf{49.3} & \textbf{53.7} & \textbf{27.5} & \textbf{30.7} & \underline{41.9} & \underline{49.5} & \textbf{44.2} & \textbf{47.8} & \textbf{48.3} & \textbf{53.1} & \textbf{31.6} & \textbf{34.4}\\

\midrule
\textsc{Mixtral-8x22b} w/ \textbf{5 psgs} & 34.5 & 38.8 & 37.0 & 46.0 & 44.1 & 47.6 & 21.3 & 24.4 & 34.4 & 41.0 & 33.9 & 36.8 & 45.0 & 49.5 & 25.9 & 28.1 \\
\textsc{Mixtral-8x22b} w/ \textbf{10 psgs} & 38.1 & 42.5 & \underline{41.8} & \underline{50.0} & \underline{47.4} & \underline{51.3} & 24.9 & 28.0 & 39.1 & 46.0 & 38.9 & 41.1 & 46.7 & 51.5 & 28.1 & 30.7 \\
\bottomrule
\end{tabular}
}
\vspace{-0.2cm}
\caption{Impact of the number of passages. Evaluation results on the entire \lfrqa~test set based on top 5 or 10 passages. W and W+T indicate win and win+tie rate against \lfrqa. \textbf{bold} and \underline{underline} indicate the best and runner-up results. * means using the answer generation prompt w/o CoT.}
\label{tab:eval-mbe-10psg}
\vspace{-0.5cm}
\end{table*}


\begin{table*}
\centering
\small
\begin{tabular}{l}
\toprule
\textbf{Answer Generation Prompt} \\
\midrule
Based on the passages, provide a helpful answer to the query. Your answer must be faithful to the content in the passages. \\ Do not use your own knowledge to answer the query. If you couldn't find any helpful information in the passages, \\ respond "I couldn't find an answer." \\
\\
Passages are inside <passage></passage> tags. Query is in the <query></query> tags.
\\
\{x.passages\} \\
\\
<query>\\
\{x.question\}\\
</query>\\
\\

First, think step by step, and put your thinking in <thinking> tags. Your thinking must be shorter than 50 words. \\ Then, provide your answer. \\

\bottomrule
\end{tabular}
\caption{Prompt for answer generation. \{x.*\} indicates a component input that can be replaced by actual data. We modify this prompt slightly to be compatible with different LLMs' input formats, but the majority of the prompt, particularly instructions, remain the same. * The last section starting with ``First, think step by step...'' is what we refer to as CoT prompt. We remove it for \textsc{GPT-4o} only.}
\label{tab:answer-gen}
\end{table*}
\clearpage


\begin{table*}[h]
\centering
\small
\begin{tabular}{l}
\toprule
\textbf{Annotation Generation Prompt} \\
\midrule
Provide a response around 100 words to the query in the <query></query> tags based on the passages. Passages are \\ inside <passage></passage> tags.
The response must incorporate all candidate answers in the <ans></ans>, and \\ you are allowed to rephrase these answers in order to make your final response natural.
The response should not \\ include any information outside passages. \\
You should cite the passage number (indices) in the format of [1], [2], [3, 4], etc. at the end of each sentence. \\
\\
\{x.passages\} \\
\\
<query>\\
\{x.question\} \\
</query> \\

\bottomrule
\end{tabular}
\caption{Prompt for \textsc{GPT-4} annotations. \{x.*\} indicates a component input that can be replaced by actual data.}
\label{tab:answer-combine}
\end{table*}
\clearpage


\begin{table*}
\centering
\small
\begin{tabular}{l}
\toprule
\textbf{Pairwise Evaluation: Instruction and Rubric} \\
\midrule
We will show you a query and a pair of answers to the query. You need to provide your preference over this pair \\ of answers. \\
\\
First, try your best to determine whether the information in an answer can help truthfully answer the query. Then rate \\ your preference based on Helpfulness and Truthfulness.\\\\
- Helpfulness: information that is helpful/relevant to answer the query. An ideal answer consists of only information \\ \hspace{1.8cm} that is helpful/relevant to answer the query. \\
- Truthfulness: information that you believe is correct to answer the query. By our definition, truthful information \\ \hspace{1.9cm} should be helpful information. If you find it difficult to determine the truthfulness of some information, \\ \hspace{1.9cm} consider it untruthful. Often time, this is due to not enough context provided in the answer. Another \\ \hspace{1.9cm} source of untruthfulness is when conflicting information presented, and the answer does not reconcile \\ \hspace{1.9cm} 
 them in a coherent way.\\
\\
<rubric>\\
Here is how you judge (in the order of importance),\\
- If one answer has all truthful information while the other has some untruthful information, prefer the all truthful one. \\
- If both have some untruthful information, prefer the one with less untruthful information.\\
- If both have all truthful information, prefer the one with more truthful or helpful information.\\
- If two answers look equally good, or it is too hard to judge using the 3 cases above, then you are our "not sure" which one \\ is better.\\
</rubric> \\
\bottomrule
\end{tabular}
\caption{Instruction and rubrics for pairwise evaluation. We use this template across all LLM evaluators. This can be considered as the "system" instruction for GPT-4 and LLama-3 models.}
\label{tab:pair-eval-instruction}
\end{table*}

\begin{table*}
\centering
\small
\begin{tabular}{l}
\toprule
\textbf{Pairwise Evaluation: Example Template} \\
\midrule
Query is in the <query></query> tags.  Answer 1 is in <answer 1></answer 1>, and Answer 2 is in \\ <answer 2></answer 2>. \\
\\
<query> \\
{x.question} \\
</query> \\
\\
<answer 1> \\
{x.response1} \\
</answer 1> \\
\\
<answer 2> \\
{x.response2} \\
</answer 2> \\
\\
Review the rubric in <rubric> tags, \\
- if you prefer <answer 1>, output 1. \\
- if you prefer <answer 2>, output 2. \\
- if you are not sure, output 0. \\
\\
First, think step by step, put your thinking in <thinking></thinking> tags. Your thinking must be shorter than 50 words. \\Then, provide your rating inside <rating></rating> tags. Remember your rating should be 0 if you are not sure, and your \\ rating must be either 0, 1, or 2. \\

\bottomrule
\end{tabular}
\caption{Example template for pairwise evaluation. This template is used for both ICL examples and the final test example.}
\label{tab:pair-eval-example}
\end{table*}

\begin{table*}
\centering
\small
\begin{tabular}{l|l}
\toprule
\textbf{Label} & 1 \\\midrule
Query & difference between publicly and publically. \\\midrule
\multirow{5}{*}{Answer 1} & Both `publicly' and `publically' bear no difference in meaning, as they are essentially alternative spellings \\ 
& of the same concept. Publicly is more widely used, but the existence of 'publically' in reputable sources like \\
& the OED means it cannot be dismissed as simply incorrect. Some opinions hold that 'publicly' is the older \\ 
& irregular form, still preached by a lot of grammars, and 'publically,' on the other hand, is the newer and \\ 
& regular form. \\\midrule
\multirow{2}{*}{Answer 2} & There is no difference in meaning between 'publicly' and 'publically'; they are alternative spellings of the \\
& same word. \\\midrule
\multirow{2}{*}{Reason} & Both <answer 1> and <answer 2> are truthful. However, <answer 1> provides more truthful information \\
& as the context to compare the two terms. Therefore, <answer 1> is better. \\
\midrule
\midrule

\textbf{Label} & 2 \\\midrule
Query & what did European/American historical cooks do with the egg whites? \\
\midrule
\multirow{2}{*}{Answer 1} & Historical European and American cooks used egg whites for making egg white omelettes and egg white \\
& pasta, as well as for stiffening clothing, similar to how starch is used today. \\
\midrule
\multirow{6}{*}{Answer 2} & Egg whites have found their place in various non-culinary applications throughout history, such as \\
& in the clarification of beer and wine, in the conservation of books through bookbinding and gilding, and in \\
& makeup as an ancient form of nail polish. \\
& They were also utilized historically as a stiffening agent for clothing, similar to how starch is used today. \\
& The culinary landscape was not left untouched, with egg whites making their way into recipes for omelettes \\
& and pastas as early as the 15th century. \\
\midrule
\multirow{2}{*}{Reason} & Both <answer 1> and <answer 2> provide several usages of egg whites for European/American cooks. \\
& <answer 2> clearly provides more options with fully explained details. Therefore, <answer 2> is better.\\
\midrule
\midrule

\textbf{Label} & 0 \\\midrule
Query & should utf-16 be considered harmful? \\\midrule
\multirow{7}{*}{Answer 1} & The question of whether UTF-16 should be considered harmful is subject to differing opinions. One \\
& perspective suggests that UTF-16 is harmful due to a historical misconception about character encoding and \\
& recommends UTF-8 as the superior choice for various text interfaces. Another viewpoint argues that \\
& UTF-16 is not inherently harmful, emphasizing its utility in specific scenarios where it serves as a \\
& compromise between simplicity and compactness. The choice between UTF-16 and other encodings like \\
& UTF-8 depends on the specific requirements of the application, such as compatibility with ASCII or \\ 
& the need to efficiently encode certain character sets.\\
\midrule
\multirow{5}{*}{Answer 2} & UTF-16 should not be considered harmful. \\
& However, contrasting views argue that UTF-16 should indeed be considered harmful. \\
& Some argue that the very reason UTF-16 exists is because some time ago there used to be a misguided belief \\
& that WideChar is going to be what UCS-4 now is. \\
& Additionally, the harmfulness of UTF-16 is tied to issues with exercising code. \\
\midrule
\multirow{2}{*}{Reason} & Both <answer 1> and <answer 2> reconcile the two conflicting views with detailed explanation. \\
& I am not sure which one is better. \\
\bottomrule
\end{tabular}
\caption{In-context examples for pairwise evaluation. Labels 1, 2, and 3 mean "answer 1 is better", "answer 2 is better" and "tie", respectively. "Reason" is a model's chain-of-thought output.}
\label{tab:pair-eval-examples}
\end{table*}
\clearpage

